%% file: main_arxiv.tex
\pdfoutput=1
    
\documentclass{article}
\usepackage{iclr2027_conference,times}

\input{math_commands.tex}

\usepackage{hyperref}
\usepackage{url}
\usepackage{booktabs}
\usepackage{tabularx}
\usepackage{multirow}
\usepackage{caption}
\usepackage{xcolor}
\usepackage{colortbl}
\usepackage{graphicx}
\usepackage{pifont}
\newcommand{\cmark}{\ding{51}}
\newcommand{\xmark}{\ding{55}}
\definecolor{rowgray}{gray}{0.94}
\usepackage{amsmath}
\usepackage{amsthm}
\usepackage{amssymb}
\newtheorem{proposition}{Proposition}

\newcommand{\nmr}{RADIANT}

\title{RADIANT: Object-Centric \\ Feed-Forward Global Illumination}

\author{Sarah Zhang\thanks{Equal contribution.}\hphantom{$^{*}$}, Zhiyang Dou\footnotemark[1]\hphantom{$^{*}$}, Minghao Guo, Benjamin Tod Jones \& Wojciech Matusik\\
  [8pt]
MIT CSAIL}

\iclrfinalcopy
\begin{document}

\maketitle
\lhead{Preprint}

\input{sections/abstract}

\input{sections/1_introduction}
\input{sections/2_related_works}
\input{sections/3_methods}
\input{sections/4_experiments}

\input{sections/5_conclusion}

\clearpage

\input{sections/statements}

\bibliography{iclr2027_conference}
\bibliographystyle{iclr2027_conference}

\clearpage
\input{sections/appendix}

\end{document}

%% file: math_commands.tex
\usepackage{amsmath,amsfonts,bm}

\def\eqref#1{equation~\ref{#1}}

\def\1{\bm{1}}

\DeclareMathAlphabet{\mathsfit}{\encodingdefault}{\sfdefault}{m}{sl}
\SetMathAlphabet{\mathsfit}{bold}{\encodingdefault}{\sfdefault}{bx}{n}

%% file: sections/abstract.tex
\begin{abstract}
Feed-forward neural renderers synthesize global illumination across novel 3D scenes without per-scene optimization or runtime ray tracing.
Existing transformer-based renderers tokenize individual triangles, so self-attention scales quadratically with mesh resolution.
Yet macroscopic radiative transport is governed by the underlying surface geometry and object arrangements rather than a particular mesh discretization; finer tessellation can thus inflate token counts without changing the underlying transport physics, motivating a discretization-agnostic and object-centric representation. We introduce \nmr{} (\textbf{RAD}iance-based \textbf{I}nter-\textbf{A}ctor \textbf{N}eural \textbf{T}ransformer), an object-centric architecture for feed-forward global illumination from point inputs.
By representing each object with a compact budget of 32 surface-patch tokens, \nmr{} decouples attention sequence length from polygon counts.
In a single network, an inter-object scene transformer models 3D radiative exchange and a view decoder renders images at target camera poses.
To capture light transport without materializing $O(N^2)$ bias matrices, invariant form-factor attention embeds geometric distances and surface orientations directly into query-key dot products as $\mathrm{SE}(3)$-invariant inner products weighted by learned area and material gains.
For primary visibility without mesh connectivity, camera rays cross-attend scene tokens guided by depth-tested point splats in a shared spherical frame.
Across a 1.03M-scene benchmark under matched model capacity, \nmr{} reaches 25\,dB tone-mapped PSNR in $5\times$ fewer training steps than RenderFormer, and achieves 32.2\,dB at 600k steps (+3.3\,dB over RenderFormer and +7.6\,dB over RenderFormer-V2). We further demonstrate the scalability of \nmr{}, which completes a $512^2$ forward pass of a 500-object scene in 0.83\,s, with computational cost scaling with the number of objects rather than polygons. We will release our code and dataset upon publication.
\end{abstract}

%% file: sections/1_introduction.tex
\section{Introduction}
\label{sec:intro}

Global illumination effects (such as soft contact shadows, diffuse interreflections, and color bleeding) provide visual cues for spatial depth and surface geometry, governed by the rendering equation \citep{kajiya1986rendering}. Classical Monte Carlo path tracing \citep{kajiya1986rendering} and radiosity \citep{goral1984radiosity} compute physically accurate light transport from surface meshes and material definitions, but tracing rays at runtime remains computationally expensive. Per-scene methods such as neural radiance fields and 3D Gaussian splatting \citep{mildenhall2020nerf, kerbl2023gaussians, hadadan2021neuralradiosity} avoid runtime path tracing by fitting scene-specific representations, but cannot generalize across new environments. Feed-forward neural renderers \citep{zeng2025renderformer, zeng2026renderformerv2} instead predict radiance across unseen scenes in a single forward pass without test-time optimization.

Existing feed-forward renderers such as RenderFormer \citep{zeng2025renderformer, zeng2026renderformerv2} tokenize explicit triangle meshes, running self-attention over individual primitives ($T$ triangles). Because full attention scales as $O(T^2)$, latency and GPU memory grow rapidly with mesh resolution. Subsequent variants introduce windowed attention \citep{zeng2026renderformerv2} or object-level pooling \citep{du2026renderformerpp}, but still require explicit triangle-based geometry and compute primitive-level attention during encoding. Point clouds, by contrast, are topology-free and interface directly with depth sensors and 3D generative models. Prior work has learned light transport on point-based scene representations for static and dynamic indoor scenes \citep{sanzenbacher2020neurallighttransport}, as a direct-to-indirect illumination operator on 3D point clouds \citep{hermosilla2019latent}, and as generalizable 3D embeddings for diffuse global illumination \citep{xu2026lighttransport}. Building on these point-based light-transport methods, we investigate an object-centric formulation that combines compact surface-patch tokens, scene-level transport, and camera-ray HDR decoding guided by mesh-free primary visibility.

We introduce \nmr{} (\textbf{RAD}iance-based \textbf{I}nter-\textbf{A}ctor \textbf{N}eural \textbf{T}ransformer), an object-centric architecture that learns feed-forward global illumination directly from point inputs. In radiosity and radiative heat transfer \citep{goral1984radiosity}, energy exchange occurs between finite surface regions and objects regardless of surface tessellation. For fixed surface geometry and materials, macroscopic light transport is the same whether a surface is represented by 50 or 50{,}000 triangles. By defining attention tokens over a compact set of surface patches per object rather than individual polygons, \nmr{} decouples transport sequence length from geometric discretization, providing a unified pipeline for meshes and point clouds.

\begin{table}[t]
\centering
\caption{\textbf{Architectural and computational comparison.} \emph{Input}: input geometric modality. \emph{Triangle attention}: extent of self-attention over individual triangles. \emph{Transport tokens} and \emph{Transport attention}: sequence length and computational complexity of global light exchange ($T$ triangles, $N$ scene entities including walls and lights). \emph{Primary visibility}: mechanism resolving camera-ray intersections. \emph{Steps to 25\,dB}: optimizer steps required to attain 25\,dB tone-mapped PSNR on the 1.03M-scene benchmark under our matched protocol.}%
\label{tab:teaser}
\footnotesize
\setlength{\tabcolsep}{2.5pt}%
\resizebox{\ifdim\width>\linewidth\linewidth\else\width\fi}{!}{%
\begin{tabular}{lccccccc}
\toprule
& & & \multicolumn{2}{c}{Transport} & & & \\
\cmidrule(lr){4-5}
Method & Input & \shortstack{Triangle\\attention} & tokens & attention & \shortstack{Primary\\visibility} & \shortstack{Per-scene\\optimization} & \shortstack{Steps to\\25\,dB} \\
\midrule
Neural Radiosity & mesh$^\dagger$ & --- & --- & --- & ray tracing & required & --- \\
RenderFormer & triangles & all pairs & $T$ & $O(T^2)$ & learned & none & 80k \\
RenderFormer-V2 & triangles & windowed & $T$ & $O(T^2)$ & learned & none & n/r %
\\
RenderFormer++ & triangles & per object & $8N$ & $O(N^2)$ & G-buffer & none & --- \\
\textbf{\nmr{} (ours)} & \textbf{point inputs} & \textbf{none} & $\boldsymbol{32N}$ & $\boldsymbol{O(N^2)}$ & \textbf{point splat} & \textbf{none} & \textbf{16k ($5\times$)} \\
\bottomrule
\end{tabular}}
\vspace{1pt}
\parbox{\linewidth}{\footnotesize $^\dagger$Its neural radiance field needs no meshing, but training and rendering trace rays against the scene geometry. 
n/r: did not reach 25\,dB within 600k steps (Tab.~\ref{tab:quality}).}%
\end{table}

\nmr{} divides computation into 3D scene transport and 2D view decoding. A point encoder first extracts 32 surface-patch tokens per object via point set abstraction, bounding global transport to $O(N^2)$ attention over $32N$ tokens for $N$ scene entities without requiring manifold meshes. An inter-object scene transformer then models radiative transfer among these patch tokens. To incorporate geometric distance and surface orientation without the quadratic memory overhead of explicit pairwise bias matrices, we introduce \emph{invariant form-factor attention} (Sec.~\ref{sec:method-inv}). Pairwise Euclidean distance and surface-facing angles enter query-key dot products as exact $\mathrm{SE}(3)$-invariant inner products, modulated by learned area and material gains. For novel views, a view decoder cross-attends camera rays to the scene tokens (Sec.~\ref{sec:method-rays}). A lightweight point-splat rasterizer supplies primary hit depth, surface normals, and materials, allowing the cross-attention layers to focus on multi-bounce interreflections, soft contact shadows, and color bleeding.

Under matched parameter capacity ($\approx$206M parameters) and an identical 1.03M-scene dataset (Tab.~\ref{tab:teaser}), \nmr{} reaches 25\,dB tone-mapped PSNR in $5\times$ fewer optimizer steps than RenderFormer (16k vs.\ 80k) and finishes 600k steps at 32.2\,dB (+3.3\,dB over RenderFormer, +7.6\,dB over RenderFormer-V2). Because sequence length scales with object count rather than polygon count ($32N$ tokens vs.\ $T$ triangles), a forward pass of \nmr{} on a synthetic 500-object scene takes 0.83\,s in a runtime-only benchmark, $8.5\times$ faster than RenderFormer with FlashAttention (Sec.~\ref{sec:exp-efficiency}); rendering quality is evaluated on scenes with up to 50 entities.

Our main contributions are:
\begin{itemize}
    \setlength{\itemsep}{1pt}\setlength{\parskip}{0pt}
    \item We introduce \nmr{}, a mesh-free transformer that renders feed-forward global illumination directly from point inputs, decoupling transport complexity from polygon tessellation ($O(N^2)$ vs.\ $O(T^2)$ for $N$ scene entities and $T$ triangles, reducing median token counts from $3{,}205$ to $320$; Tab.~\ref{tab:datastats}).
    \item We formulate invariant form-factor attention, exactly factorizing selected geometric invariants motivated by radiative transfer into query-key dot products without pairwise bias tables, paired with a rasterizer-guided view decoder that resolves primary visibility without mesh connectivity.
    \item On our 1.03M-scene benchmark under matched training budgets, \nmr{} reaches 25\,dB in $5\times$ fewer optimizer steps than RenderFormer at the same global batch size, improves final fidelity by $+3.3$\,dB, and runs a forward pass on 500-object scenes in 0.83\,s in a runtime-only benchmark.
\end{itemize}

%% file: sections/2_related_works.tex
\section{Related Work}
\label{sec:related}

\noindent \textbf{Per-scene neural rendering.}
Neural radiance fields \citep{mildenhall2020nerf, mueller2022instantngp}, 3D Gaussian splatting \citep{kerbl2023gaussians}, and Neural Radiosity \citep{hadadan2021neuralradiosity} optimize continuous representations for a single scene by minimizing photometric loss or the rendering equation's residual \citep{kajiya1986rendering}. Although these methods avoid path tracing once trained, they require per-scene fitting and cannot generalize to novel geometric arrangements. In contrast, our goal is feed-forward synthesis: predicting global illumination across novel scenes in a single forward pass without test-time optimization.

\noindent \textbf{Feed-forward rendering of explicit geometry.}
RenderFormer \citep{zeng2025renderformer} showed that transformers \citep{vaswani2017attention} can learn light transport over triangle meshes by tokenizing triangles with their material attributes. RenderFormer-V2 \citep{zeng2026renderformerv2} adds environment lighting and participating media with windowed attention, but still relies on triangle primitives. Concurrently, RenderFormer++ \citep{du2026renderformerpp} pools triangle tokens into object queries, yet still executes full self-attention over each object's triangles before pooling, which incurs $O(\sum_i T_i^2)$ cost. RenderFormer++ is concurrent and has not released code; it requires triangle-based geometry and triangle-level attention, whereas \nmr{} operates directly on point inputs via set abstraction \citep{qi2017pointnetpp} without computing primitive attention. Other object-centric representations model scattering functions \citep{guo2020osf}, object-oriented transfer \citep{zheng2023nelt}, superposed fields \citep{zheng2024superposed}, or light probes \citep{deng2025transgi}. In our framework, surface patches represent the physical scale of macroscopic light exchange directly.

\noindent \textbf{Point-based and alternative representations for global illumination.}
Meshless representations have long decoupled illumination from surface tessellation \citep{lehtinen2008meshless}. \citet{hermosilla2019latent} learn a direct-to-indirect illumination operator on 3D point clouds and propagate its features to image pixels, while \citet{xu2026lighttransport} map point clouds to continuous 3D shading queries for diffuse transport. Related approaches aggregate virtual point lights via pixel-light cross-attention \citep{ren2024lightformer} or predict screen-space shading from G-buffers \citep{nalbach2017deepshading}. Here, \nmr{} pools point inputs into 32 surface-patch tokens per object, models multi-bounce transfer in 3D self-attention, and decodes HDR radiance along camera rays guided by a point-splat G-buffer (Appendix~\ref{app:related}).

\noindent \textbf{Rotary embeddings and attention biases.}
Spatial geometry and surface orientation determine radiative transfer. Rotary position embeddings (RoPE) \citep{su2024roformer} encode relative offsets into query-key dot products, maintaining compatibility with fused attention kernels \citep{dao2022flashattention}. PRoPE \citep{li2025prope} extends rotary embeddings to camera frustums for view synthesis, from which we adopt the CamRay parameterization. Rather than storing explicit $O(N^2)$ pairwise bias matrices, low-rank factorizations \citep{wu2025flashbias} decompose spatial biases into inner products. Invariant form-factor attention adopts this principle: Euclidean distance and surface-facing angles enter as exact inner products of per-token vectors, weighted by gains predicted from surface area and material attributes. We additionally incorporate feature-wise linear modulation (FiLM) \citep{perez2018film} and register tokens \citep{darcet2024registers} to condition on shape statistics and prevent attention sinks.

%% file: sections/3_methods.tex
\section{Method}
\label{sec:method}
\noindent
\textbf{Problem setup.}
A scene is defined as a set of $N$ objects $\{\mathcal{O}_i\}_{i=1}^{N}$, each specified by surface geometry with spatially varying material attributes (diffuse albedo, specular reflectance, roughness, and emission; emissive objects serve as light sources), together with a query camera with intrinsics $K$ and pose $[R\,|\,t]$. The goal is to synthesize, in a single forward pass for novel scenes, the HDR radiance image $L \in \mathbb{R}^{H\times W\times 3}$ that a reference path tracer would produce. We represent each object by $P{=}2{,}048$ surface point samples, each carrying a 16-dimensional attribute vector comprising position, normal, and ten material parameters. \nmr{} is natively \emph{mesh-free}. Polygonal meshes undergo uniform surface sampling, while raw point clouds with normals and materials are ingested directly without surface reconstruction. Our primary benchmarks evaluate mesh-derived samples, while Sec.~\ref{sec:exp-quality} and Appendix~\ref{app:partial} also evaluate simulated single-view partial point clouds.

\subsection{Architecture}
\label{sec:method-overview}

\begin{figure}[t]
\centering
\includegraphics[width=\linewidth]{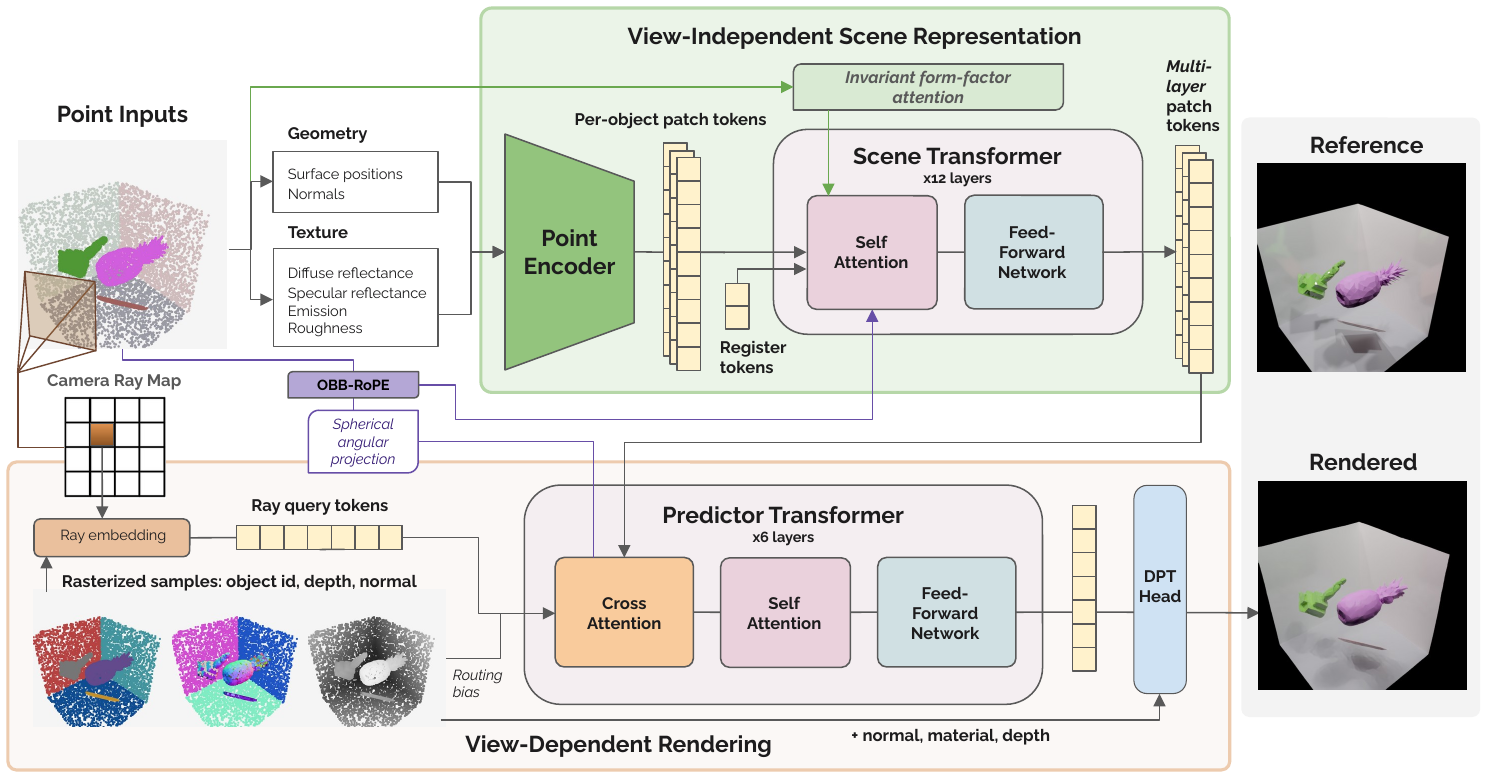}
\caption{\textbf{Overview of \nmr{}}, a single feed-forward network. View-independent part (computed once per scene): the point inputs are pooled into 32 patch tokens per object, and a scene transformer resolves 3D light transport with invariant form-factor attention.
View-dependent part (computed per query viewpoint): the view decoder (predictor transformer), guided by rasterized geometry channels and spherical RoPE, cross-attends camera rays to scene tokens to decode the HDR image.}
\label{fig:pipeline}
\end{figure}
Fig.~\ref{fig:pipeline} illustrates the network architecture. First, a point encoder maps each object's point inputs to $M{=}32$ surface-patch tokens (Sec.~\ref{sec:method-tokens}). Global radiative transfer is modeled by a \emph{scene transformer} (12 self-attention layers, width 1{,}024, 8 heads, feed-forward width 1{,}890, 97.0M parameters) with rotary encodings of object pose (Sec.~\ref{sec:method-tokens}) and invariant form-factor attention (Sec.~\ref{sec:method-inv}). Next, for each viewpoint, the view decoder (or \emph{predictor}, 6 layers with alternating cross- and self-attention, 96.1M parameters) renders the image by cross-attending camera ray tokens (constructed from CamRay geometry and point-splat rasterization; Sec.~\ref{sec:method-rays}) to the scene tokens. Finally, a dense prediction transformer (DPT) head \citep{ranftl2021dpt} upsamples the decoded ray tokens from $4\times4$ pixel patches to full-resolution HDR radiance.

\subsection{Object tokenization}
\label{sec:method-tokens}

A PointNet-style encoder \citep{qi2017pointnet} (10.0M parameters) embeds $P$ point inputs per object in the object's centered, scale-normalized coordinate frame. Using PointNet++ set abstraction \citep{qi2017pointnetpp}, farthest-point sampling (FPS) selects $M{=}32$ centroids per object with 128 nearest neighbors, where local max-pooling and linear projection yield 32 patch tokens of width 1{,}024 per object. Sixteen learnable register tokens \citep{darcet2024registers} absorb extraneous attention weight ($-0.13 \pm 0.01$\,dB when removed; Tab.~\ref{tab:ablation-full}). By construction, the scene transformer is equivariant to object permutations, ensuring that rendered imagery is invariant to object input ordering (Prop.~\ref{prop:perm}, Appendix~\ref{app:proofs}).

To express relative 3D geometry, we inject object pose into attention queries and keys via an object-centric rotary embedding (\textit{OBB-RoPE}). Each token is assigned a 12-dimensional oriented bounding box (OBB) descriptor $\mathbf{x}^{(i)} \in \mathbb{R}^{12}$, concatenating the patch centroid with the object's three principal axes scaled by their standard deviations (computed in world space for the scene transformer and camera space for the predictor; Appendix~\ref{app:arch}).\label{sec:method-obbrope} 
The rotations of queries and keys compose into a function $R(\mathbf{x}^{(j)} - \mathbf{x}^{(i)})$ of the relative descriptor (Prop.~\ref{prop:rel}).

\subsection{Radiosity statistics and conditioning}
\label{sec:method-bias}

Each object carries eight radiosity statistics shared by its tokens: a log area proxy (1), mean diffuse albedo (3), mean emission (3), and scalar mean specular reflectance (1). Learned receiver and sender encodings $\delta\mathbf{q}_i$ and $\delta\mathbf{k}_j$ of these statistics are added to the queries and keys of the scene transformer before rotary encoding (Appendix~\ref{app:arch}), and the same statistics determine the gains of invariant form-factor attention (Sec.~\ref{sec:method-inv}). Two FiLM pathways~\citep{perez2018film} (2.1M parameters each) further condition each object's tokens on rotation-invariant shape statistics of its point inputs and, in the view-dependent part, on its pose relative to the query camera.
\label{sec:method-film}

\subsection{Invariant form-factor attention}
\label{sec:method-inv}

Although rotary position embeddings capture relative positions, they do not directly expose the geometric invariants that govern radiative exchange, namely pairwise Euclidean distance $r_{ij}$ and the angle $\theta_i$ between the receiver normal and the inter-patch displacement vector $\boldsymbol{\Delta}_{ij} = \mathbf{c}_j - \mathbf{c}_i$, and the angle $\theta_j$ between the sender normal and $-\boldsymbol{\Delta}_{ij}$. Patch centroids are centered at the scene mean and normalized by its root-mean-square radius (Prop.~\ref{prop:inv}).
We formulate an exact inner-product factorization that embeds these $O(3)$-invariant geometric terms directly into query-key dot products. Each patch token carries its centroid $\mathbf{c}_i$ and unnormalized mean normal $\mathbf{m}_i$ (averaged over the 128 point inputs of its patch). We construct a 4-dimensional geometric invariant vector $\mathbf{G}_{ij} \in \mathbb{R}^4$ capturing the primary terms of radiative transfer (embedded via a 16-dimensional factored query-key representation):
\begin{equation}
\mathbf{G}_{ij} = \Big( 
\underbrace{-\|\boldsymbol{\Delta}_{ij}\|^2}_{\text{distance kernel}}, \;\; 
\underbrace{\mathbf{m}_i^{\top}\boldsymbol{\Delta}_{ij}}_{\text{receiver-facing}}, \;\; 
\underbrace{-\mathbf{m}_j^{\top}\boldsymbol{\Delta}_{ij}}_{\text{sender-facing}}, \;\; 
\underbrace{\mathbf{m}_i^{\top}\mathbf{m}_j}_{\text{mutual orientation}} 
\Big).
\label{eq:inv}
\end{equation}
Each of these four invariant scalar terms expands as an exact inner product between receiver and sender features. For example, the negative squared distance expands as $-\|\boldsymbol{\Delta}_{ij}\|^2 = \langle [2\mathbf{c}_i, -\|\mathbf{c}_i\|^2, -1], [\mathbf{c}_j, 1, \|\mathbf{c}_j\|^2] \rangle$. Appending these factored features directly to the scene transformer queries and keys injects a learned form-factor prior without materializing a pairwise bias matrix. Across layers and attention heads, these terms are modulated by receiver and sender gains $a_t(i)$ and $b_t(j)$ predicted from radiosity statistics (Sec.~\ref{sec:method-bias}). This contributes a learned scalar bias $\sum_{t=1}^4 a_t(i) b_t(j) G_{ij,t}$ to each attention logit without materializing the scene-space geometric bias matrix. Further, this bias is a material-dependent function of distance and orientation; when the product of the distance gains is positive, its squared-distance term contributes a Gaussian-shaped multiplicative factor to the unnormalized attention weights.

\noindent \textbf{Theoretical properties and scope.}
These terms provide the exact $O(3)$-invariant quadratic arguments of the differential radiative form factor $dF_{ij} = \cos\theta_i\cos\theta_j\,dA_j/(\pi r_{ij}^2)$ \citep{goral1984radiosity} (Prop.~\ref{prop:inv}). Non-quadratic physical phenomena (specifically non-negative cosine clamping $[\cos\theta]_+$, inverse-square distance attenuation $1/r_{ij}^2$, and mutual visibility occlusion) are not explicitly represented by this factorization and are left to the learned transformer layers. All main experiments (Sec.~\ref{sec:exp-quality}) evaluate \nmr{} with this formulation, and we isolate and measure its contribution in the ablation study (Sec.~\ref{sec:exp-ablations}).

\subsection{View-dependent decoding: ray tokens, rasterizer channels, and spherical cross-attention}
\label{sec:method-rays}

\noindent \textbf{Ray tokens.} 
We tokenize camera rays by adopting an unnormalized variant of the CamRay parameterization \citep{li2025prope}. Each pixel's \emph{CamRay}, obtained from $K^{-1}[u_{\mathrm{pix}}, v_{\mathrm{pix}}, 1]^{\top}$ and converted to the OpenGL camera frame, preserves the linearity of the image plane. Pixels are grouped into $4\times4$ patches. For each patch, the CamRay map and per-pixel channels described below are linearly projected and layer-normalized into a single ray token positioned at the patch's CamRay centroid.

\noindent \textbf{Rasterizer channels.} 
To avoid burdening cross-attention with geometric ray casting, we separate primary surface visibility from global light transport. A depth-tested point-splat rasterization of the point inputs for the query camera ($7\times7$ dilation) provides a mesh-free primary G-buffer with visible object identity, depth, normal, and material per pixel. The buffer does not explicitly encode incident illumination or shadows; cross-attention to the 3D scene tokens supplies global context for predicting direct and indirect illumination. Besides reflectance, the buffer's material channels carry only the emission of directly visible emitters. The resulting log-depth, camera-space normal, and hit mask are patchified alongside the CamRay map, ensuring each ray token carries its primary hit geometry. Object identities enter cross-attention as a \emph{routing bias}: the attention logit between a ray patch and an object's tokens receives an additive bias proportional to the fraction of patch pixels rasterized to that object, scaled by a learned per-layer signed gate (initialized to zero). For output generation, rasterized normal, materials, and depth (14 channels) are added to full-resolution DPT features via a zero-initialized $1\times1$ convolution.

\noindent \textbf{Spherical cross-attention RoPE.} 
Cross-attention requires comparing 3D object positions against 2D image-plane coordinates under a unified relative encoding. We project both into a spherical coordinate frame consisting of two angles in $[-\pi, \pi]$ and a depth coordinate: an object at camera-space position $(X, Y, Z)$ maps to $(\operatorname{atan2}(X, -Z), \operatorname{atan2}(Y, -Z), -Z)$, whereas a ray with normalized image-plane coordinates $(u, v)$, i.e., OpenGL camera-space direction proportional to $(u, v, -1)$, maps to $(\arctan u, \arctan v, \bar d)$, where $\bar d$ denotes the mean rasterized hit depth of the patch (or $0$ for missed rays). Under RoPE subtraction, the difference in the first two coordinates directly measures the angular offset between the ray and the object from the camera viewpoint, while the third coordinate provides explicit depth information for occlusion reasoning. These angular coordinates coincide exactly when a ray points toward the object's centroid (Prop.~\ref{prop:sph}).

\subsection{Training objective}
\label{sec:method-training}

The network is trained end-to-end. Following \citet{zeng2025renderformer}, the model predicts log radiance $\hat{y} = \log(1 + L)$. The loss $\mathcal{L}$ combines mean absolute error in log space with a perceptual term \citep{zhang2018lpips} on tone-mapped images $\tau(L) = \operatorname{clip}_{[0,1]}(\log_2(1 + L))$:
\begin{equation}
\mathcal{L} = \big\| \hat{y} - \log(1 + L^{\star}) \big\|_1 + 0.05\, \mathrm{LPIPS}\big(\tau(\hat{L}), \tau(L^{\star})\big),
\end{equation}
where the first term is averaged over all pixels and color channels, $L^{\star}$ denotes ground-truth radiance, and $\hat{L} = \exp(\min(\hat{y}, 10)) - 1$. We optimize using AdamW \citep{loshchilov2019adamw} with a peak learning rate of $5\times10^{-5}$, 8{,}000 warmup steps, cosine learning rate decay to $10^{-6}$, and a global batch size of 24. All models share this loss, tone map, LPIPS backbone, and optimizer setup, except where RenderFormer-V2's released training loop differs (Appendix~\ref{app:training}).

%% file: sections/4_experiments.tex
\section{Experiments}
\label{sec:experiments}

\subsection{Experimental setup}
\label{sec:exp-setup}

\noindent \textbf{Data.}
Following the RenderFormer protocol \citep{zeng2025renderformer}, we construct procedural scenes with 1--6 Objaverse objects \citep{deitke2023objaverse} drawn from 18{,}085 meshes in a template room with randomized wall pruning, area lights (1--8 emitters per scene, mean 4.5; Tab.~\ref{tab:datastats}), and per-shading-group materials. Four views per scene are rendered with Blender Cycles at $512^2$ resolution (512 samples per pixel, denoised via Intel Open Image Denoise; Appendix~\ref{app:data}). All models are trained from scratch on our 1.03M-scene corpus without external pre-rendered data. Apart from the frozen LPIPS backbone in the loss, the only pretrained components are the frozen encoders in RenderFormer-V2's released input pipeline (Appendix~\ref{app:arch}). The held-out evaluation split comprises 1{,}028 scenes (4{,}112 images) featuring novel object compositions, materials, emitters, and camera viewpoints. Each model ingests identical scenes in its native format (surface point samples for \nmr{}, triangle meshes for baselines).

\begin{table}[t]
\centering
\caption{Held-out rendering quality under the matched-budget protocol ($512^2$, 4{,}112 images). Each model is evaluated at its best checkpoint, selected over all evaluated checkpoints on the same split (Appendix~\ref{app:training}). Bold indicates best performance.}
\label{tab:quality}
\small\setlength{\tabcolsep}{8pt}
\begin{tabular}{lccc}
\toprule
Model & PSNR $\uparrow$ (per-image / pooled) & LPIPS $\downarrow$ & $\ell_1(\log)$ $\downarrow$ \\
\midrule
\nmr{} (ours) & \textbf{32.22} / \textbf{27.65} & \textbf{0.042} & \textbf{0.012} \\
RenderFormer & 28.95 / 25.80 & 0.077 & 0.019 \\
RenderFormer-V2 & 24.60 / 23.05 & 0.117 & 0.046 \\
\bottomrule
\end{tabular}
\end{table}

\noindent \textbf{Matched-budget protocol.}
We benchmark three models of matched parameter capacity: \nmr{} (205.9M), RenderFormer-Base \citep{zeng2025renderformer} (205.2M), and RenderFormer-V2 (207.6M; Appendix~\ref{app:arch}), retrained under our unified protocol at $512^2$ resolution on NVIDIA A100 GPUs (batch size 24, AdamW, bfloat16 mixed precision with FP32 master weights; Sec.~\ref{sec:method-training}). Training runs for 600k steps, starting with a 100k-step schedule and extended across successive cosine decays with periodic checkpoint evaluations (Sec.~\ref{sec:exp-quality}, Appendix~\ref{app:training}). Under the $256^2$ protocol, seed-to-seed standard deviation is $0.01$\,dB, and paired bootstrap 95\% confidence intervals confirm that our $+3.3$\,dB and $+7.6$\,dB performance margins are statistically significant (Appendix~\ref{app:seeds}).

\noindent \textbf{Metrics.}
Models are evaluated on the 4{,}112 held-out images using three complementary metrics: (1) PSNR following Reinhard tone mapping \citep{reinhard2002tonemapping} and sRGB conversion, computed both \emph{per-image} and \emph{pooled} across all pixels prior to logarithmic compression, (2) LPIPS \citep{zhang2018lpips} with an AlexNet backbone, and (3) mean absolute error in log radiance $\ell_1(\log)$ (the regression term of the loss).

\subsection{Rendering quality}
\label{sec:exp-quality}

\begin{figure}[t]
\centering
\begin{minipage}[c]{0.68\linewidth}
\includegraphics[width=\linewidth]{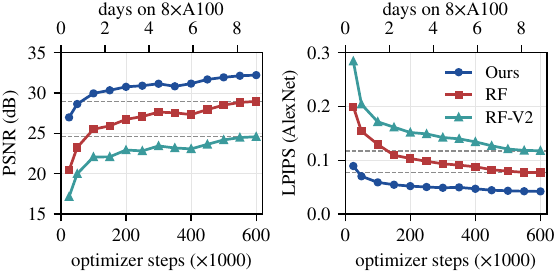}
\end{minipage}\hfill
\begin{minipage}[c]{0.3\linewidth}
\captionsetup{justification=raggedright,singlelinecheck=false,skip=0pt}
\caption{Learning curves of \nmr{} (Ours), RenderFormer (RF), and RenderFormer-V2 (RF-V2) on 4{,}112 held-out images: per-image PSNR (left) and LPIPS (right). Top axis: 8$\times$A100 days at 1.28\,s/step, the median step time of our model's training run. Labels in \nmr{}-equivalent time, not per-baseline wall-clock time. Dashed: best values of RF and RF-V2 (Tab.~\ref{tab:quality}).}
\label{fig:curves}
\end{minipage}
\end{figure}

\nmr{} consistently outperforms both RenderFormer baselines across training, converging in substantially fewer optimizer steps. Learning curves (Fig.~\ref{fig:curves}) and evaluation checkpoints (Tab.~\ref{tab:quality-full}) show that by step 6k, \nmr{} leads RenderFormer by $+7.30$\,dB (23.45 vs.\ 16.15\,dB) and cuts perceptual error by more than half (LPIPS 0.156 vs.\ 0.350). By step 16k, \nmr{} reaches 25.91\,dB, surpassing RenderFormer at step 50k (23.25\,dB) and reaching 25\,dB in five times fewer optimizer steps at the same global batch size (16k vs.\ 80k).

Upon completing the 600k-step schedule, \nmr{} achieves 32.21\,dB per-image PSNR (27.65\,dB pooled), outperforming RenderFormer by $+3.26$\,dB ($+1.85$\,dB pooled) and RenderFormer-V2 by $+7.63$\,dB, while reducing LPIPS by 45\% and 64\%, respectively (0.042 vs.\ 0.077 and 0.117; Tab.~\ref{tab:quality-full}). Across training, the margin over RenderFormer remains substantial (+5.4\,dB at 50k, +4.4\,dB at 100k, +3.5\,dB at 300k, and +3.3\,dB at 600k). RenderFormer-V2, retrained with its authors' released implementation under our matched budget, trails both models across all checkpoints. Across all evaluated checkpoints on the held-out split (Tab.~\ref{tab:quality}; Appendix~\ref{app:training}), the best checkpoints are \nmr{} at 596k (32.22\,dB), RenderFormer at 600k (28.95\,dB), and RenderFormer-V2 at 599k (24.60\,dB).

\begin{table}[t]
\centering
\footnotesize
\caption{Held-out quality by scene complexity: the 4{,}112 evaluation images grouped by the number of entities the
renderer sees (foreground objects, walls, and lights; the corpus spans 3 to 18). Per-image PSNR (dB) and LPIPS
(AlexNet), with all three models at matched step 600k. The three models' per-image scores are aligned by image index, which
identifies the same procedural scene across models (Appendix~\ref{app:data}).}
\label{tab:objcount}
\resizebox{\ifdim\width>\linewidth\linewidth\else\width\fi}{!}{%
\begin{tabular}{lrcccccc}
\toprule
& & \multicolumn{2}{c}{\nmr{} (ours)} & \multicolumn{2}{c}{RenderFormer} & \multicolumn{2}{c}{RenderFormer-V2} \\
\cmidrule(lr){3-4}\cmidrule(lr){5-6}\cmidrule(lr){7-8}
Entities per scene & Images & PSNR $\uparrow$ & LPIPS $\downarrow$ & PSNR $\uparrow$ & LPIPS $\downarrow$ & PSNR $\uparrow$ & LPIPS $\downarrow$ \\
\midrule
$\leq 7$ & 796 & \textbf{32.35} & \textbf{0.037} & 29.22 & 0.064 & 24.46 & 0.104 \\
8--10 & 1{,}376 & \textbf{32.15} & \textbf{0.041} & 29.00 & 0.075 & 24.52 & 0.116 \\
11--13 & 1{,}336 & \textbf{32.44} & \textbf{0.043} & 29.11 & 0.079 & 24.80 & 0.118 \\
$\geq 14$ & 604 & \textbf{31.64} & \textbf{0.051} & 28.12 & 0.094 & 24.35 & 0.136 \\
\bottomrule
\end{tabular}}
\end{table}

\begin{table}[t]
\centering
\footnotesize
\caption{Unseen assets and larger scenes (Appendix~\ref{app:unseen}): per-image PSNR / LPIPS of the step-600k checkpoints; four views per scene. The held-out row repeats the 600k evaluation of Tab.~\ref{tab:quality-full}. Bold: best per row.}
\label{tab:unseen}
\begin{tabular}{lcc ccc}
\toprule
Test set & Scenes & Entities & \nmr{} (ours) & RenderFormer & RenderFormer-V2 \\
\midrule
Held-out split & 1{,}028 & 3--18 & \textbf{32.21} / \textbf{0.042} & 28.95 / 0.077 & 24.57 / 0.117 \\
\emph{A}: unseen assets & 250 & 3--17 & \textbf{32.59} / \textbf{0.041} & 29.36 / 0.073 & 24.90 / 0.110 \\
\emph{B}: 25--50 entities & 280 & 25--50 & \textbf{25.32} / \textbf{0.095} & 22.50 / 0.171 & 20.06 / 0.227 \\
\emph{C}: object sizes of \emph{B} & 250 & 3--18 & \textbf{34.95} / \textbf{0.024} & 32.35 / 0.044 & 26.44 / 0.079 \\
\bottomrule
\end{tabular}
\end{table}

\nmr{} maintains a $3.1$--$3.5$\,dB margin over RenderFormer across all entity counts (Tab.~\ref{tab:objcount}; Appendix~\ref{app:qual}). On an independent test set of 500 novel Objaverse assets (Tab.~\ref{tab:unseen}; Appendix~\ref{app:unseen}), \nmr{} still leads RenderFormer and RenderFormer-V2 by $+3.2$\,dB and $+7.7$\,dB at training scene scales, and by $+2.8$\,dB and $+5.3$\,dB on crowded scenes with 25--50 entities.

\begin{figure}[p]
\centering
\includegraphics[width=\linewidth]{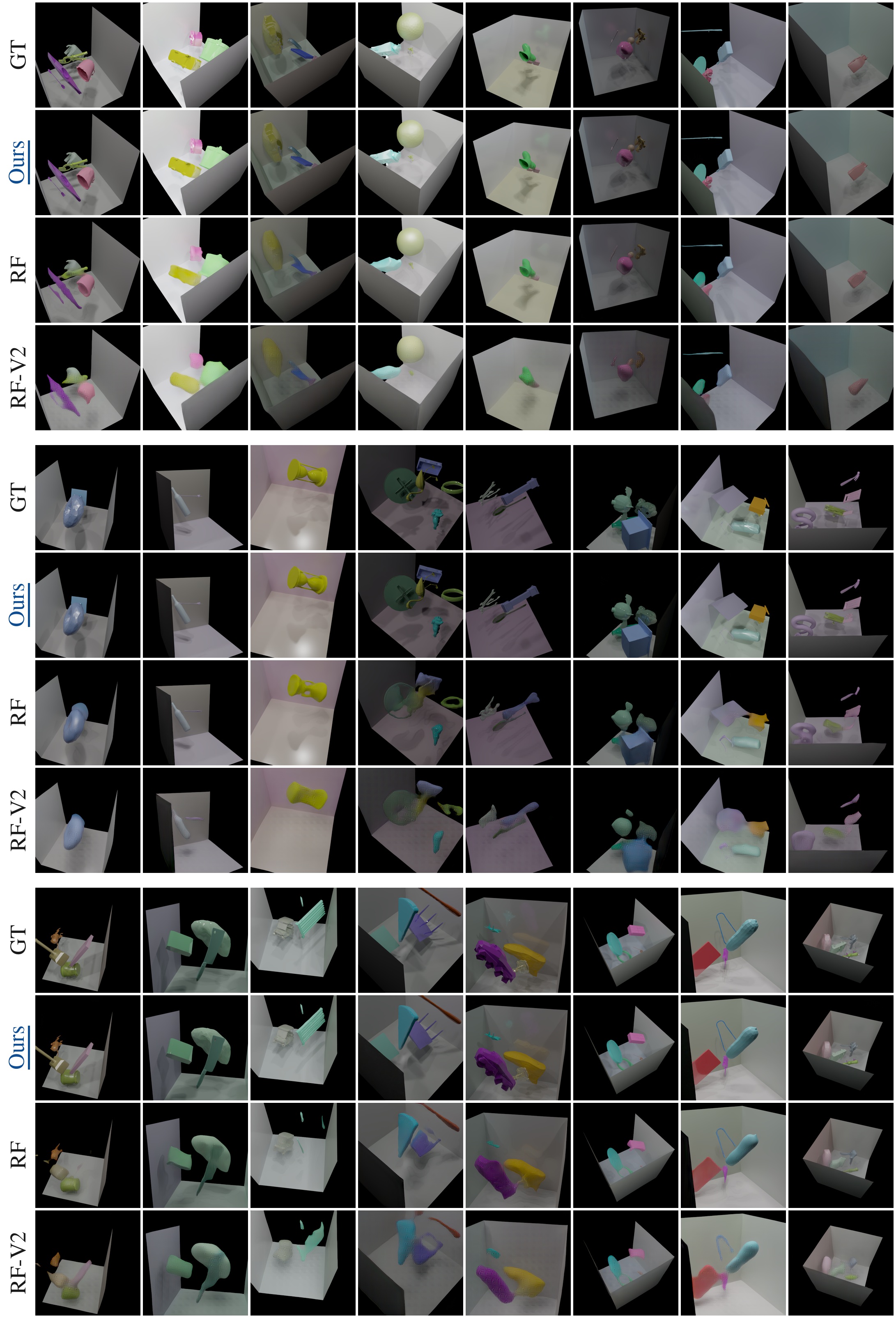}
\caption{Qualitative comparisons on 24 crowded held-out scenes at matched step 600k. Rows in each group of eight scenes: ground truth (GT), \nmr{} (Ours), RenderFormer (RF), and RenderFormer-V2 (RF-V2).}%
\label{fig:qual}
\end{figure}

On crowded held-out scenes (Fig.~\ref{fig:qual}, step 600k), \nmr{} resolves contact shadows, fine silhouettes, and color bleeding between nearby surfaces. In contrast, both RenderFormer baselines exhibit blurring and blocking artifacts across planar and curved geometry (detailed error breakdown in Appendix~\ref{app:qual}).

\begin{figure}[t]
\vspace{0mm}
\centering
\includegraphics[width=\linewidth]{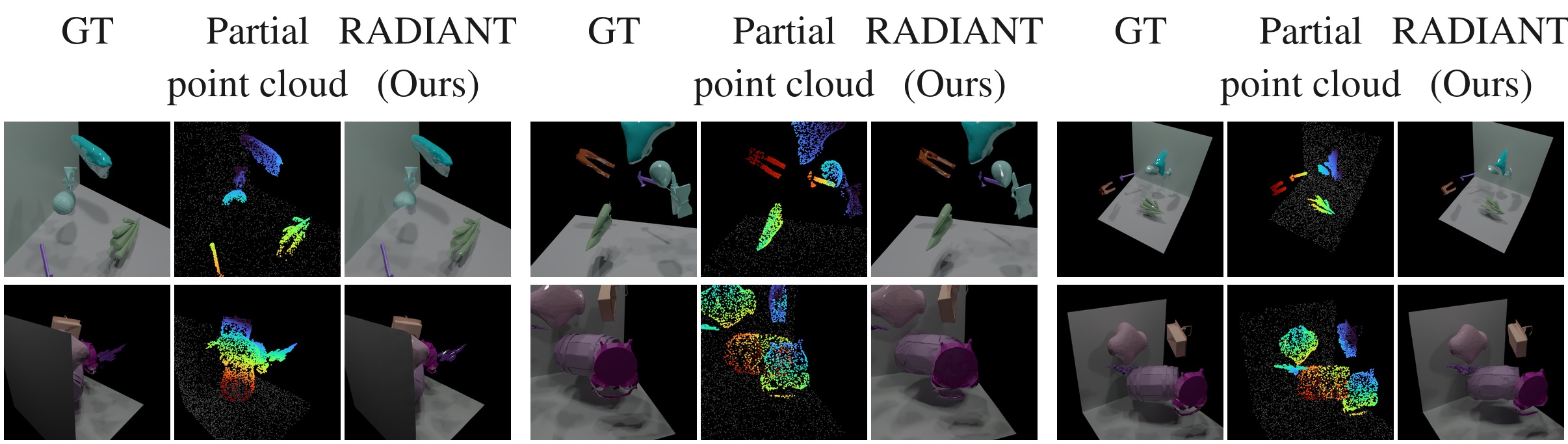}
\caption{Rendering from partial point clouds (Appendix~\ref{app:partial}) on two held-out scenes (rows), three novel views each, none used for fine-tuning. Each group of three columns: ground truth, the simulated single-view partial point cloud, and \nmr{} after 400 fine-tuning steps on four reference views.}
\label{fig:partial}
\end{figure}

The mesh-free formulation also handles incomplete point inputs without surface meshing. On novel viewpoints, \nmr{} achieves 26.33\,dB zero-shot from simulated single-view partial point clouds missing 52\%--72\% of each foreground object's point inputs, compared to 34.24\,dB from complete geometry (Tab.~\ref{tab:partial}). Fine-tuning for 400 steps on four reference views raises novel-view PSNR to 30.30\,dB, recovering half of the missing-geometry gap (Fig.~\ref{fig:partial}, Appendix~\ref{app:partial}).

\subsection{Ablations}
\label{sec:exp-ablations}

\begin{table}[t]
\centering
\footnotesize
\caption{Rendering from simulated single-view scans: the two held-out scenes of Fig.~\ref{fig:partial}, \nmr{} at step 600k (per-image PSNR, LPIPS, and
$\ell_1$ in log radiance; the evaluator of Sec.~\ref{sec:exp-setup}). \emph{Original views}: the four views of each scene, which
supervise the fine-tuning (eight images); \emph{new views}: eight further views per scene from the training camera distribution, never
used for fitting (16 images).}
\label{tab:partial}
\setlength{\tabcolsep}{3.5pt}
\begin{tabular}{llcccccc}
\toprule
& & \multicolumn{3}{c}{Original views} & \multicolumn{3}{c}{New views} \\
\cmidrule(lr){3-5}\cmidrule(lr){6-8}
Point inputs & Fine-tuning & PSNR $\uparrow$ & LPIPS $\downarrow$ & $\ell_1(\log)$ $\downarrow$ & PSNR $\uparrow$ & LPIPS $\downarrow$ & $\ell_1(\log)$ $\downarrow$ \\
\midrule
Complete & none & 33.87 & 0.034 & 0.004 & 34.24 & 0.037 & 0.004 \\
Partial (single-view scan) & none & 26.76 & 0.085 & 0.010 & 26.33 & 0.100 & 0.010 \\
Partial (single-view scan) & 400 steps & 37.51 & 0.015 & 0.003 & 30.30 & 0.065 & 0.006 \\
Complete & 400 steps & 39.02 & 0.009 & 0.002 & 33.58 & 0.038 & 0.005 \\
\bottomrule
\end{tabular}
\end{table}

\begin{table}[t]
\centering
\caption{Ablation of key mechanisms ($256^2$, 100k steps, mean over seeds 42--44; $^\dagger$seeds 42--47). PSNR per-image (mean $\pm$ sample standard deviation) / pooled; $\Delta$: per-image change from the full model, paired within each seed ($^\dagger$seeds 45--47 against the full model's three-seed mean). Extended ablations in Tab.~\ref{tab:ablation-full}.}
\label{tab:ablation}
\resizebox{\ifdim\width>\linewidth\linewidth\else\width\fi}{!}{%
\footnotesize
\setlength{\tabcolsep}{5pt}%
\begin{tabular}{>{\raggedright\arraybackslash}p{5.3cm}cccc}
\toprule
Variant & PSNR $\uparrow$ & LPIPS $\downarrow$ & $\ell_1(\log)$ $\downarrow$ & $\Delta$ \\
\midrule
Full model & $\mathbf{29.96} \pm 0.01$ / \textbf{26.16} & \textbf{0.059} & \textbf{0.019} & --- \\
No invariant form-factor attention (\S\ref{sec:method-inv}) & $29.43 \pm 0.07$ / 25.92 & 0.063 & 0.021 & $-0.52 \pm 0.08$ \\
No inter-object scene attention (\S\ref{sec:method-overview})$^\dagger$ & $27.50 \pm 0.66$ / 24.89 & 0.072 & 0.036 & $-2.46 \pm 0.66$ \\
No rasterizer channels (hit geometry, routing bias) & $26.84 \pm 0.07$ / 24.41 & 0.092 & 0.025 & $-3.12 \pm 0.08$ \\
\bottomrule
\end{tabular}}
\end{table}

Tab.~\ref{tab:ablation} evaluates individual architectural components under a controlled 100k-step schedule at $256^2$ resolution across 2{,}948 held-out images.

\noindent \textbf{3D light transport modeling.}
Restricting scene attention to intra-object tokens degrades PSNR by $2.46 \pm 0.66$\,dB across six random seeds and nearly doubles log-radiance error from $0.019$ to $0.036$ (Appendix~\ref{app:seeds}): the view decoder's cross-attention to each object cannot replace global radiative exchange within the scene transformer. Removing invariant form-factor attention (Sec.~\ref{sec:method-inv}) costs $0.52 \pm 0.08$\,dB and raises LPIPS by 7\%. A precision control confirms that float32 attention alone does not account for this gain: setting the invariant geometric terms to zero while retaining float32 attention degrades quality by $0.53 \pm 0.02$\,dB (Tab.~\ref{tab:ablation-full}).

\noindent \textbf{Explicit visibility grounding.}
Removing the rasterizer channels (hit geometry and routing bias) lowers PSNR by $3.12 \pm 0.08$\,dB and increases LPIPS by 56\% (Tab.~\ref{tab:ablation}), even though these channels carry only unshaded geometry (depth, normal, and hit mask). This shows providing explicit primary surface hits relieves cross-attention from having to reconstruct visibility. Conversely, on the model without invariant form-factor attention, decoding the rasterized G-buffer without 3D scene tokens lowers PSNR by $5.66 \pm 0.06$\,dB (Tab.~\ref{tab:raster}), demonstrating the importance of global 3D scene context beyond the view-local rasterized buffer.

\noindent \textbf{Token efficiency and auxiliary modules.}
On the model without invariant form-factor attention, varying the patch token budget eightfold ($M \in \{8, 16, 32, 64\}$) alters PSNR by at most $0.04$\,dB (Tab.~\ref{tab:tokens}): allocating 16 patch tokens achieves $29.47$\,dB (matching 32 tokens at $29.43$\,dB) while reducing training step time by 17\%, showing that compact patch sets reliably capture macroscopic transport. Conversely, pooling RenderFormer's triangle tokens into 32 tokens per object lowers its PSNR by $0.38 \pm 0.09$\,dB (Appendix~\ref{app:seeds}), indicating that architectural formulation rather than token budget accounts for \nmr{}'s performance. Removing FiLM conditioning lowers PSNR by $0.15 \pm 0.02$\,dB, while removing register tokens costs $0.13 \pm 0.01$\,dB (Tab.~\ref{tab:ablation-full}). 

\subsection{Inference efficiency and scaling with scene complexity}
\label{sec:exp-efficiency}

Because \nmr{}'s sequence length scales with object count ($32N$) rather than surface primitives, scene self-attention requires $O(N^2)$ operations regardless of geometric tessellation. On small scenes (up to 20 objects at $512^2$), RenderFormer is faster because \nmr{}'s runtime is dominated by decoding 16{,}384 ray tokens. Beyond 25 objects (35 with FlashAttention), \nmr{} becomes faster. In a runtime-only benchmark with random weights on synthetic 500-object scenes, \nmr{} runs $8.5\times$ faster than RenderFormer with FlashAttention and $22.2\times$ faster than standard PyTorch SDPA (benchmarks across tessellation levels, memory footprints, training costs, and throughput in Appendix~\ref{app:runtime}). In crowded regimes with 25--50 entities beyond the training distribution, rendering quality drops for all evaluated models (Tab.~\ref{tab:unseen}).

%% file: sections/5_conclusion.tex
\section{Conclusion}
\label{sec:conclusion}

\nmr{} shows that feed-forward global illumination can be learned directly from point inputs without mesh connectivity or per-scene optimization. By grouping point geometry into compact object-centric surface patches, the architecture decouples attention sequence length from polygon counts, while invariant form-factor attention factorizes geometric invariants of radiative transfer into query-key inner products without explicit pairwise bias tables. On a 1.03M-scene benchmark, \nmr{} trains to 25\,dB in one-fifth the steps of RenderFormer, achieves 32.2\,dB at 600k steps, and, in a runtime-only benchmark with 500 objects, runs $8.5\times$ faster than RenderFormer.

\paragraph{Limitations and future work.}
Our formulation uses a uniform budget of 32 patch tokens per object. While effective for typical assets, this choice can over-allocate tokens to simple planar walls and under-sample geometrically complex objects. Allocating tokens adaptively based on surface area or curvature is left to future work. In addition, full self-attention over ray tokens in the view decoder limits scaling beyond $1024^2$ resolution, which windowed or hierarchical ray attention could alleviate. Our point-splat rasterizer currently handles only direct surface intersections, leaving volumetric media such as smoke or fog to future work. Finally, high-frequency noise in input normals can propagate into rendered planar surfaces (Tab.~\ref{tab:robust}). Pre-filtering normals with learned or bilateral smoothing could improve robustness on noisy real-world scans.

\label{end:main}

%% file: sections/statements.tex
\subsection*{AI use statement}\label{sec:ai-statement}

Large language models (LLMs) were used to polish the manuscript prose and to assist with figure visualization.

\subsection*{Ethics statement}

This work uses only synthetic scenes assembled from publicly available Objaverse assets \citep{deitke2023objaverse} and involves no human subjects or personal data. All assets carry Creative Commons licenses. Of the 18{,}085 training meshes, 16{,}573 are CC BY, 739 CC BY-NC, 378 CC BY-SA, 354 CC BY-NC-SA, and 41 CC0; of the 500 meshes of the unseen-asset test set (Appendix~\ref{app:unseen}), 474 are CC BY, 14 CC BY-NC-SA, 8 CC BY-NC, and 4 CC BY-SA. The released scene manifest will list the license and author of every asset, and the non-commercial and share-alike terms of these licenses apply to the scenes that contain such assets.

\subsection*{Reproducibility statement}

The matched-budget protocol (Sec.~\ref{sec:exp-setup}) and Appendices~\ref{app:training}--\ref{app:arch} specify model architectures, parameter counts, dataset generation, loss functions, optimizer configurations, and evaluation schedules for all models. All models are trained under a matched configuration (Appendix~\ref{app:training} lists the details of RenderFormer-V2's released training loop that differ) and scored with identical metric definitions. We will release our complete data-generation pipeline, scene corpus, training configurations, and evaluation code.

%% file: sections/appendix.tex
\appendix
\section{Error analysis and scene complexity}
\label{app:qual}
\noindent \textbf{Remaining errors.} In Fig.~\ref{fig:qual}, the errors of \nmr{} concentrate on thin structures, which the rasterizer's $7\times7$ dilation renders slightly thicker than in the reference, and on the boundary between an object and its cast shadow, where the hit channels and the routing bias change discontinuously.

\noindent \textbf{Performance breakdown by scene complexity.} The gap measured over the whole held-out split is consistent across subsets of
scenes. Tab.~\ref{tab:objcount} splits the same 4{,}112 images by the number of entities the renderer sees, from the sparsest bucket (at most 7 entities, 796 images) to the most crowded. At matched step 600k, \nmr{}'s lead over RenderFormer is $3.1$--$3.5$\,dB in every bucket, and its lead over RenderFormer-V2 is $7.3$--$7.9$\,dB, confirming that the performance hierarchy of Tab.~\ref{tab:quality} is not an artifact of scene complexity. Absolute PSNR varies by at most 1.1\,dB with the entity count for all three models, which is
expected: the corpus holds one to six foreground objects in a template room, so the buckets differ in how much of the
image is occupied rather than in the difficulty of the transport. The perceptual metric does separate them: from the
sparsest to the most crowded bucket LPIPS rises by 37\% for \nmr{}, 47\% for RenderFormer, and 31\% for RenderFormer-V2; the fixed budget of surface-patch tokens per object does not make crowded scenes disproportionately harder. This is a statement about rendering quality only, as the inference cost of \nmr{} and RenderFormer at large object counts is measured separately in Tab.~\ref{tab:scaling}.

\section{Properties of the geometric mechanisms}
\label{app:proofs}
\subsection{Statements}
\begin{proposition}[Object-permutation equivariance]
\label{prop:perm}
The scene transformer is equivariant to permutations of the object set, and the predictor's output is invariant to them: for any permutation $\pi$ of $\{1,\dots,N\}$, applying $\pi$ to the objects permutes the scene-transformer token blocks by $\pi$ and, provided the rasterization selects the same surface at every pixel under relabeling (no depth ties between different objects), leaves the rendered image unchanged.
\end{proposition}
\begin{proposition}[Relative-pose dependence]
\label{prop:rel}
For the rotary block, the attention logit between patch tokens $i$ and $j$ depends on $\mathbf{x}^{(i)}$ and $\mathbf{x}^{(j)}$ only through their difference: $\tilde{\mathbf{q}}_i^{\top}\tilde{\mathbf{k}}_j = \mathbf{q}_i^{\top} R(\mathbf{x}^{(j)} - \mathbf{x}^{(i)})\, \mathbf{k}_j$ for a block-diagonal rotation $R$ determined by the frequencies, for fixed pre-rotation vectors $\mathbf{q}_i, \mathbf{k}_j$. In particular, a common translation of all centroids leaves the rotary-block logits between patch tokens unchanged.
\end{proposition}
\begin{proposition}[Angular alignment]
\label{prop:sph}
For a ray with normalized image-plane coordinates $(u,v)$ and an object at camera-space position $(X,Y,Z)$ with $Z<0$, the first two spherical coordinates coincide exactly when the ray passes through the object's centroid, and their difference is the pair of view angles between the ray and the centroid direction. The angular coordinates are therefore invariant to reparameterizations of the intrinsics and resolution that leave the physical rays unchanged.
\end{proposition}
\begin{proposition}[Invariant form-factor attention]
\label{prop:inv}
Let $\mathbf{x}_i = (\mathbf{c}_i - \bar{\mathbf{c}})/\rho$ be the token centroids normalized per scene ($\bar{\mathbf{c}}$ the mean and $\rho$ the root-mean-square radius of the scene's patch centroids), let $\mathbf{G}_{ij}$ be as in Eq.~\ref{eq:inv} with $\boldsymbol{\Delta}_{ij} = \mathbf{x}_j - \mathbf{x}_i$, and let $\ell_{ij} = \sum_t a_t(i)\, b_t(j)\, G_{ij,t}$ be the geometric term of the logit.
\begin{enumerate}
    \item Each $G_{ij,t}$ equals $\langle \boldsymbol{\phi}_t(i), \boldsymbol{\psi}_t(j) \rangle$ for per-token features $\boldsymbol{\phi}_t, \boldsymbol{\psi}_t$ of total width 16.
    \item $\ell_{ij}$ is unchanged when every centroid is mapped to $R\mathbf{c} + \mathbf{t}$ and every normal to $R\mathbf{m}$, for any $R \in O(3)$ and $\mathbf{t} \in \mathbb{R}^3$. $\mathbf{G}_{ij}$ is moreover unchanged under a uniform scaling $\mathbf{c} \mapsto s\mathbf{c}$, $s > 0$, and $\ell$ permutes with the tokens.
    \item Every $O(3)$-invariant polynomial in $(\boldsymbol{\Delta}_{ij}, \mathbf{m}_i, \mathbf{m}_j)$ that is homogeneous of degree two is a linear combination of the four entries of $\mathbf{G}_{ij}$, $\|\mathbf{m}_i\|^2$, and $\|\mathbf{m}_j\|^2$.
\end{enumerate}
\end{proposition}
\subsection{Proofs}
\begin{proof}[Proof of Proposition~\ref{prop:perm}]
The encoder of Sec.~\ref{sec:method-tokens} is applied independently to each object, so permuting the objects permutes its outputs. The OBB descriptor $\mathbf{x}^{(i)}$, the radiosity encodings $\delta\mathbf{q}_i,\delta\mathbf{k}_i$, and the FiLM statistics are each functions of object $i$ alone, so they permute with the tokens; the register tokens' position is the mean patch centroid, which is permutation-invariant. Self-attention with per-token positional terms is permutation-equivariant, hence so is the scene transformer. In the predictor, ray tokens attend to the set of patch tokens through a softmax over the set, which is permutation-invariant. The rasterizer channels depend on the union of the objects' surfaces and, under the hypothesis that no two objects tie for the nearest depth at a pixel, select the same surface after relabeling. The routing bias attaches to each object's tokens by identity. The rendered image is therefore invariant.
\end{proof}
\begin{proof}[Proof of Proposition~\ref{prop:rel}]
Write the rotary block as $\tilde{\mathbf{q}}_i = R(\mathbf{x}^{(i)})\mathbf{q}_i$ and $\tilde{\mathbf{k}}_j = R(\mathbf{x}^{(j)})\mathbf{k}_j$, where $R(\mathbf{x})$ is block-diagonal with $2\times2$ rotations by angles $\omega_f x_d$ for coordinate $d$ and frequency $\omega_f$. Then $\tilde{\mathbf{q}}_i^{\top}\tilde{\mathbf{k}}_j = \mathbf{q}_i^{\top} R(\mathbf{x}^{(i)})^{\top} R(\mathbf{x}^{(j)}) \mathbf{k}_j = \mathbf{q}_i^{\top} R(\mathbf{x}^{(j)} - \mathbf{x}^{(i)}) \mathbf{k}_j$ because planar rotations commute and compose additively. A common translation $\mathbf{t}$ of all centroids adds $\mathbf{t}$ to the first three coordinates of every patch descriptor and leaves their differences, hence the rotary-block logits between patch tokens, unchanged. The content vectors are translation-invariant because the encoder sees centered samples and the FiLM and radiosity statistics are translation-invariant.
\end{proof}
\begin{proof}[Proof of Proposition~\ref{prop:sph}]
A ray with normalized image-plane coordinates $(u,v)$ has camera-space direction proportional to $(u, v, -1)$, so its horizontal and vertical view angles are $\arctan u$ and $\arctan v$. An object centroid at $(X,Y,Z)$ with $Z<0$ has view angles $\operatorname{atan2}(X,-Z)$ and $\operatorname{atan2}(Y,-Z)$. The two pairs coincide exactly when $(X,Y)/(-Z) = (u,v)$, i.e., when the ray passes through the centroid, and their differences are the angular offsets between the ray and the centroid direction. Both quantities are defined from camera-space geometry and the corresponding unnormalized camera-space ray, so a change of focal length or resolution that leaves the physical ray unchanged leaves the coordinates unchanged.
\end{proof}
\begin{proof}[Proof of Proposition~\ref{prop:inv}]
1. By expansion,
\begin{align*}
-\|\mathbf{x}_j - \mathbf{x}_i\|^2 &= \langle [2\mathbf{x}_i, -\|\mathbf{x}_i\|^2, -1], [\mathbf{x}_j, 1, \|\mathbf{x}_j\|^2] \rangle, \\
\mathbf{m}_i^{\top}(\mathbf{x}_j - \mathbf{x}_i) &= \langle [\mathbf{m}_i, -\mathbf{m}_i^{\top}\mathbf{x}_i], [\mathbf{x}_j, 1] \rangle, \\
-\mathbf{m}_j^{\top}(\mathbf{x}_j - \mathbf{x}_i) &= \langle [\mathbf{x}_i, 1], [\mathbf{m}_j, -\mathbf{m}_j^{\top}\mathbf{x}_j] \rangle, \\
\mathbf{m}_i^{\top}\mathbf{m}_j &= \langle \mathbf{m}_i, \mathbf{m}_j \rangle,
\end{align*}
of widths $5 + 4 + 4 + 3 = 16$; scaling the query features of term $t$ by $a_t(i)$ and the key features by $b_t(j)$ makes $\ell_{ij}$ their inner product. 

2. Under $\mathbf{c} \mapsto R\mathbf{c} + \mathbf{t}$ the mean maps to $R\bar{\mathbf{c}} + \mathbf{t}$ and $\rho$ is unchanged, so $\mathbf{x} \mapsto R\mathbf{x}$; with $\mathbf{m} \mapsto R\mathbf{m}$, every entry of $\mathbf{G}_{ij}$ is an inner product of $\boldsymbol{\Delta}_{ij}$, $\mathbf{m}_i$, $\mathbf{m}_j$ and is preserved because $R^{\top}R = I$. Under $\mathbf{c} \mapsto s\mathbf{c}$, $\rho$ scales by $s$, so $\mathbf{x}$ is unchanged, and the normals do not change. The gains are functions of per-object statistics (log area, albedo, emission, specular reflectance) that are invariant to rigid motions, though the log area is not invariant to scaling. Hence $\ell_{ij}$ is invariant to rigid motions. Every quantity is a function of its own token and of permutation-invariant scene statistics ($\bar{\mathbf{c}}$, $\rho$), so $\ell$ permutes with the tokens. 

3. By the first fundamental theorem of invariant theory for $O(3)$ \citep{weyl1946classical}, used for equivariant networks by \citet{villar2021scalars}, every $O(3)$-invariant polynomial in vectors $\mathbf{v}_1, \dots, \mathbf{v}_k$ is a polynomial in their inner products $\langle \mathbf{v}_a, \mathbf{v}_b \rangle$. Those homogeneous of degree two are linear combinations of the inner products themselves, which for $(\boldsymbol{\Delta}_{ij}, \mathbf{m}_i, \mathbf{m}_j)$ are the four entries of $\mathbf{G}_{ij}$ up to sign, $\|\mathbf{m}_i\|^2$, and $\|\mathbf{m}_j\|^2$. The last two depend on one token alone: a query-only term cancels in the softmax, and a key-only term is a per-key offset.
\end{proof}

\section{Training details}
\label{app:training}
Tab.~\ref{tab:recipe} summarizes the training configuration and hyperparameters in our experiments. All models are trained from scratch at $512^2$ on NVIDIA A100 GPUs with a global batch size of 24 (without data augmentation). We optimize using AdamW \citep{loshchilov2019adamw} with $\beta_1 = 0.9$, $\beta_2 = 0.999$, weight decay of $0.01$ applied to matrix weights, and gradient clipping at a global norm of 1.0. To train RenderFormer, we use its released model code, Base configuration, and training initialization within the training loop shared with \nmr{}. RenderFormer-V2 is trained also with its released training loop, with the data loader replaced to read our scenes and the LPIPS backbone set to AlexNet--this loop applies the weight decay to all parameters, warms up from zero, anneals to $5\times10^{-7}$, and casts the targets to bfloat16. Both baselines keep their view-dependent stage in TF32, as in their authors' code. 

The learning rate warms up linearly over 8{,}000 steps to a peak of $5\times 10^{-5}$, followed by cosine annealing to a minimum of $10^{-6}$ at 100k steps. Longer runs continue from the 100k checkpoint on the tail of a 200k-step cosine schedule, from the 200k checkpoint on that of a 300k-step one, and from the 300k checkpoint on that of a 600k-step one (same warm-up, peak, and minimum; model and optimizer state carried over), so the learning rate rises at each continuation, for example to $2.6\times 10^{-5}$ at step 300k; the main comparison is at step 600k. Training uses bfloat16 mixed precision with FP32 master weights. 

Every saved checkpoint is evaluated on the full held-out split of 4{,}112 images with identical metric definitions (one evaluator for \nmr{} and RenderFormer, a port of it to RenderFormer-V2's released code); the best checkpoint of each model (Tab.~\ref{tab:quality}) is the one with the highest per-image PSNR on this split, selected identically for the three models. Choosing each model's checkpoint on a random half of the held-out scenes and scoring it on the other half (1{,}000 halvings) selects the 600k checkpoint in every halving for both RenderFormer generations and in 99.9\% of the choices for \nmr{} (otherwise its 575k checkpoint) among the regular evaluations from 100k steps on, and the best checkpoints of Tab.~\ref{tab:quality} exceed these choices by at most $0.02$\,dB.
\begin{table}[htbp]
\centering
\caption{Training hyperparameters of the main comparison. The ablations (Sec.~\ref{sec:exp-ablations}) use $256^2$ images, 100k steps, seeds 42--44 (42--47 without inter-object attention), and 2{,}948 held-out images.}
\label{tab:recipe}
\footnotesize
\setlength{\tabcolsep}{3.5pt}
\setlength{\aboverulesep}{0pt}
\setlength{\belowrulesep}{0pt}
\renewcommand{\arraystretch}{1.25}
\begin{tabular}{>{\raggedright\arraybackslash}p{2.6cm}>{\raggedright\arraybackslash}p{3.7cm}@{\hspace{1.2em}}>{\raggedright\arraybackslash}p{2.6cm}>{\raggedright\arraybackslash}p{3.7cm}}
\toprule
\textbf{Hyperparameter} & \textbf{Value} & \textbf{Hyperparameter} & \textbf{Value} \\
\midrule
Hardware & NVIDIA A100 GPUs & Gradient clipping & Global norm 1.0 \\
\rowcolor{rowgray} Global batch size & 24 & Mixed precision & bf16 autocast, fp32 master weights \\
Optimizer & AdamW, $(\beta_1,\beta_2)=(0.9, 0.999)$ & Loss function & MAE in $\log(1{+}L) + 0.05\,\text{LPIPS}$ \\
\rowcolor{rowgray} Peak learning rate & $5\times10^{-5}$ & LPIPS backbone & AlexNet \\
Warm-up & 8{,}000 linear steps & Image resolution & $512 \times 512$ \\
\rowcolor{rowgray} Learning rate decay & Cosine to $10^{-6}$ at 100k/200k/300k/600k & Data augmentation & None (1 view per sample) \\
Weight decay & 0.01 & Evaluation split & Held-out (4{,}112 images) \\
\rowcolor{rowgray} Random seed & 42 (initialization and data order; RenderFormer-V2's released loop sets it after building the model) & Total training steps & 100k, continued to 600k on longer schedules; matched comparison at 600k \\
\bottomrule
\end{tabular}
\end{table}

\begin{table}[htbp]
\centering
\caption{Evaluation points of Fig.~\ref{fig:curves} (step 25k and every 50k) and the early steps quoted in Sec.~\ref{sec:exp-quality} (6k, 16k, 80k): per-image PSNR (dB) and LPIPS (AlexNet) on the 4{,}112 held-out images. ---: not evaluated. Bold: best value of each run.}
\label{tab:quality-full}
\small
\begin{tabular}{@{}r@{\hspace{12pt}}cc@{\hspace{12pt}}cc@{\hspace{12pt}}cc@{}}
\toprule
 & \multicolumn{2}{c}{\nmr{} (ours)} & \multicolumn{2}{c}{RenderFormer} & \multicolumn{2}{c}{RenderFormer-V2} \\
\cmidrule(lr){2-3}\cmidrule(lr){4-5}\cmidrule(lr){6-7}
Step & PSNR $\uparrow$ & LPIPS $\downarrow$ & PSNR $\uparrow$ & LPIPS $\downarrow$ & PSNR $\uparrow$ & LPIPS $\downarrow$ \\
\midrule
6k & 23.45 & 0.156 & 16.15 & 0.350 & --- & --- \\
16k & 25.91 & 0.103 & 19.02 & 0.244 & 16.39 & 0.326 \\
25k & 26.97 & 0.089 & 20.42 & 0.199 & 17.13 & 0.284 \\
50k & 28.63 & 0.070 & 23.25 & 0.154 & 19.98 & 0.204 \\
80k & 29.70 & 0.061 & 25.06 & 0.133 & 21.75 & 0.175 \\
100k & 29.95 & 0.059 & 25.50 & 0.129 & 22.06 & 0.171 \\
150k & 30.33 & 0.054 & 25.92 & 0.109 & 22.08 & 0.161 \\
200k & 30.76 & 0.052 & 26.72 & 0.103 & 22.93 & 0.152 \\
250k & 30.90 & 0.050 & 27.08 & 0.098 & 22.81 & 0.149 \\
300k & 31.14 & 0.049 & 27.62 & 0.093 & 23.43 & 0.142 \\
350k & 30.82 & 0.050 & 27.54 & 0.091 & 23.18 & 0.140 \\
400k & 31.16 & 0.047 & 27.31 & 0.088 & 23.07 & 0.134 \\
450k & 31.67 & 0.044 & 27.94 & 0.082 & 23.63 & 0.127 \\
500k & 31.93 & 0.043 & 28.49 & 0.079 & 24.18 & 0.121 \\
550k & 32.13 & \textbf{0.042} & 28.83 & \textbf{0.077} & 24.47 & 0.118 \\
600k & \textbf{32.21} & \textbf{0.042} & \textbf{28.95} & \textbf{0.077} & \textbf{24.57} & \textbf{0.117} \\
\bottomrule
\end{tabular}
\end{table}

\section{Data statistics}
\label{app:data}
Tab.~\ref{tab:datastats} summarizes the corpus. Scenes follow the RenderFormer protocol: one to six foreground objects from a pool of 18{,}085 curated Objaverse meshes, placed in a template room whose walls are randomly pruned, with randomized area lights, materials assigned per shading group, and meshes simplified by quadric edge collapse. Each scene is path traced in Blender Cycles from four cameras at $512^2$ with 512 samples per pixel, and a second corpus generated by the same procedure is rendered at $256^2$ with 256 samples per pixel. Both corpora use the Cycles defaults of Blender~4.2: adaptive sampling (noise threshold 0.01), an indirect-light clamp of 10, and denoising with Intel Open Image Denoise. Every entity the renderer sees (foreground objects, walls, and lights) is one object of the model, which is why the entity count exceeds the foreground count. 
The held-out split is the last 0.1\% of scenes in a fixed order, and we verify that the point-sampled and triangle representations carry identical scene sets in identical order and that the split boundary falls on the same scene. The held-out scenes share the curated pool of 18{,}085 Objaverse meshes with the training set, but they are distinct scenes: their compositions, arrangements, material assignments, emitters, and camera viewpoints are sampled independently of the training scenes. Per-scene statistics are measured on a random sample of 40{,}000 training scenes (40 data files of 1{,}000 scenes).
\begin{table}[htbp]
\centering
\footnotesize
\caption{Corpus statistics. Entity, light, and triangle statistics are from a random sample of 40{,}000 training scenes (40 data files of 1{,}000 scenes); p5/p50/p95 are percentiles.}
\label{tab:datastats}
\resizebox{\ifdim\width>\linewidth\linewidth\else\width\fi}{!}{%
\begin{tabular}{lr}
\toprule
Scenes (after exclusions) & 1{,}027{,}596 \\
Images per scene / total images & 4 / 4{,}110{,}384 \\
Resolutions / samples per pixel & $512^2$ / 512 and $256^2$ / 256 \\
Training / held-out scenes & 1{,}026{,}568 / 1{,}028 (4{,}112 images) \\
Foreground objects per scene & 1--6 (from 18{,}085 meshes) \\
Entities per scene (objects, walls, lights) & mean 10.1; p5/p50/p95 = 5/10/15; max 18 \\
Emissive entities per scene & mean 4.5; range 1--8 \\
Triangles per scene (simplified) & mean 3{,}196; p5/p50/p95 = 1{,}155/3{,}205/5{,}630; cap 8{,}192 \\
Point inputs per object (\nmr{}) & 2{,}048, with normals and 10 material parameters \\
Excluded scenes & 322 (154 cap, 119 archive, 48 timeout, 1 remesh) \\
\bottomrule
\end{tabular}}
\end{table}

\section{Runtime analysis}
\label{app:runtime}
\noindent \textbf{Complexity.} With $N$ objects, $M{=}32$ patch tokens per object, and 16 register tokens, the scene transformer attends over $S = 32N + 16$ tokens at cost $O(S^2)$ per layer; the predictor attends $Q = HW/16$ ray tokens over the scene tokens at cost $O(QS)$ and over themselves at cost $O(Q^2)$ ($Q = 16{,}384$ at $512^2$); the rasterization is one splat pass over the $NP$ point inputs ($P = 2{,}048$) with a $7\times7$ dilation; the DPT head is linear in $HW$. None of these terms depends on the triangle count of the input meshes. A triangle-level model attends over $T$ tokens at cost $O(T^2)$ in its first stage.

\begin{figure}[htbp]
\centering
\includegraphics[width=\linewidth]{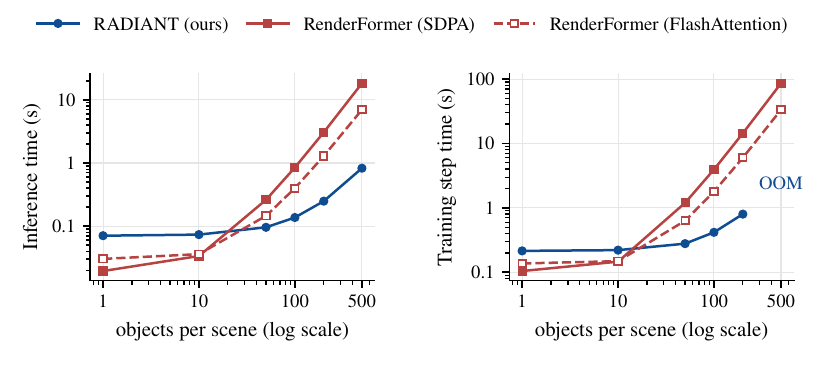}
\caption{Inference forward latency (left) and training step latency (right) of the two 206M models versus the number of objects ($512^2$, batch size 1, bf16 autocast, one H100 80\,GB GPU; 512 triangles per object for RenderFormer, 2{,}048 point inputs per object for \nmr{}). RenderFormer is evaluated with PyTorch SDPA (solid) and FlashAttention (dashed). Training step includes forward pass, loss computation, backward pass, and AdamW update. In complex scenes with 50--500 objects, \nmr{} provides up to $22.2\times$ faster inference than RenderFormer with PyTorch SDPA and $8.5\times$ faster than RenderFormer with FlashAttention. Training steps of \nmr{} are $4.3$--$18.0\times$ faster than RenderFormer's with SDPA and $2.3$--$7.5\times$ faster than with FlashAttention at 50--200 objects; at 500 objects \nmr{} exceeds the 80\,GB of GPU memory in training, whereas RenderFormer needs 32.7\,GiB. Inference latencies are tabulated in Tab.~\ref{tab:scaling}.}
\label{fig:efficiency}
\end{figure}

\noindent \textbf{Protocol.} Tabs.~\ref{tab:scaling} and~\ref{tab:scaling-all}: the two 206M models of Tab.~\ref{tab:quality} with random weights, synthetic scenes (2{,}048 point inputs per object sampled on spheres, with analytic normals, for \nmr{}; random small triangles for RenderFormer; surface sampling is not timed), bf16 autocast, batch size 1, one NVIDIA H100 GPU (80\,GB; PyTorch 2.7.1, CUDA 12.6), five warm-up and up to twenty timed forward passes per configuration (median reported). Tab.~\ref{tab:throughput}: the same models evaluated on eight-object scenes (512 triangles per object for RenderFormer, 2{,}048 points per object for \nmr{}), with five warm-up and twenty timed passes, excluding host-to-device transfers. Fig.~\ref{fig:efficiency} plots the latencies of Tab.~\ref{tab:scaling} against the number of objects, together with the latency of a training step. Complete benchmarking scripts and profiling configurations are included in the supplementary materials.

\noindent \textbf{Breakdown of the forward pass.} Tab.~\ref{tab:stages} splits the forward pass into its components. The view-dependent part dominates \nmr{} at 10 and 50 objects (68\% and 58\% of the time) and grows only from 50 to 75\,ms up to 200 objects, while the point encoder and the scene transformer, whose attention runs in float32 (Sec.~\ref{sec:method-inv}), grow with the number of objects (7.3 to 48.0\,ms and 9.0 to 104.0\,ms). RenderFormer's triangle transformer grows from 19\,ms to 1.20\,s with FlashAttention over the same range and takes 94\% of the forward pass at 200 objects (96\% with PyTorch attention).
\begin{table}[htbp]
\centering
\caption{Inference time per component (ms, median of 20 forwards, CUDA events around the top-level modules) of the two 206M models at $512^2$ on one H100 with random weights, batch size 1, and bf16, for 10, 50, and 200 objects (2{,}048 point inputs or 512 triangles per object). Encoder: point encoder of \nmr{} or triangle attribute encodings of RenderFormer; transport: scene transformer or triangle transformer; view decoder + DPT: view-dependent transformer and DPT head; other: rasterization, conditioning, and the remaining operations. RenderFormer with FlashAttention (flash) and, at 200 objects, PyTorch attention (sdpa). Profiled separately from Tab.~\ref{tab:scaling}.}
\label{tab:stages}
\footnotesize
\setlength{\tabcolsep}{4pt}
\begin{tabular}{lrrrrrrr}
\toprule
 & \multicolumn{3}{c}{\nmr{}} & \multicolumn{4}{c}{RenderFormer} \\
\cmidrule(lr){2-4}\cmidrule(lr){5-8}
Component & 10 & 50 & 200 & 10 (flash) & 50 (flash) & 200 (flash) & 200 (sdpa) \\
\midrule
Encoder & 7.3 & 16.0 & 48.0 & 0.3 & 0.9 & 3.0 & 3.0 \\
Transport & 9.0 & 16.0 & 104.0 & 19.2 & 116.1 & 1{,}203.0 & 2{,}959.9 \\
View decoder + DPT & 50.2 & 55.7 & 75.4 & 15.6 & 26.8 & 71.4 & 101.5 \\
Other & 7.5 & 7.7 & 20.6 & 1.2 & 1.4 & 2.6 & 2.9 \\
\midrule
Total & 74.0 & 95.4 & 248.0 & 36.3 & 145.3 & 1{,}280.0 & 3{,}067.3 \\
\bottomrule
\end{tabular}
\end{table}
\begin{table}[htbp]
\centering
\caption{Single-forward inference time (ms, median) and peak memory (MiB) vs.\ number of objects ($512^2$, batch size 1, H100), from 2{,}048 point inputs per object for \nmr{} and 512 triangles per object for RenderFormer (surface sampling not timed). RenderFormer runs with PyTorch SDPA or FlashAttention (its memory is that of the SDPA run); speedups divide its time by \nmr{}'s.}
\label{tab:scaling}
\footnotesize
\begin{tabular}{lrrrrrrr}
\toprule
& \multicolumn{2}{c}{\nmr{}} & \multicolumn{3}{c}{RenderFormer} & \multicolumn{2}{c}{Speedup} \\
\cmidrule(lr){2-3}\cmidrule(lr){4-6}\cmidrule(lr){7-8}
Objects (triangles) & time & memory & sdpa & flash & memory & sdpa & flash \\
\midrule
1 (512)         & 70.4  & 4{,}255  & 19.4       & 30.2      & 1{,}631  & $0.28\times$ & $0.43\times$ \\
10 (5{,}120)    & 73.3  & 4{,}299  & 33.6       & 35.9      & 1{,}644  & $0.46\times$ & $0.49\times$ \\
50 (25{,}600)   & 95.7  & 4{,}480  & 260.8      & 145.4     & 1{,}954  & $2.7\times$  & $1.5\times$  \\
100 (51{,}200)  & 136.8 & 5{,}863  & 848.1      & 398.5     & 2{,}859  & $6.2\times$  & $2.9\times$  \\
200 (102{,}400) & 248.5 & 10{,}898 & 3{,}065.2  & 1{,}288.9 & 4{,}669  & $12.3\times$ & $5.2\times$  \\
500 (256{,}000) & \textbf{826.9} & 25{,}996 & 18{,}326.0 & 6{,}990.0 & 10{,}101 & $22.2\times$ & $8.5\times$ \\
\bottomrule
\end{tabular}
\end{table}

\noindent \textbf{Crossover.} Measured between 10 and 50 objects with the settings of Tab.~\ref{tab:scaling}, RenderFormer with SDPA is faster at $512^2$ up to 20 objects (70.9 versus 77.2\,ms at 20), and \nmr{} is faster from 25 objects on; against FlashAttention the crossover lies between 30 and 35 objects. At $128^2$ and $256^2$, \nmr{} is faster from 15 objects on (at $128^2$ from 10 against FlashAttention).

\noindent \textbf{Tessellation.} Tab.~\ref{tab:tess} keeps ten objects and varies only their tessellation. RenderFormer's time grows with the triangle count, while \nmr{}'s input (2{,}048 points per object) and time do not change: from 2{,}048 triangles per object on, \nmr{} is faster ($1.5$--$2.5\times$), and at 32{,}768 triangles per object by $155\times$ (FlashAttention) to $412\times$ (SDPA).
\begin{table}[htbp]
\centering
\caption{Single-forward inference time (ms, median) for ten objects as their tessellation grows ($512^2$, batch size 1, H100, settings of Tab.~\ref{tab:scaling}). \nmr{} reads 2{,}048 point inputs per object regardless of tessellation. Speedups divide RenderFormer's time by \nmr{}'s.}
\label{tab:tess}
\footnotesize
\begin{tabular}{lrrrrr}
\toprule
Triangles per object (total) & \nmr{} & RF sdpa & RF flash & Speedup (sdpa) & Speedup (flash) \\
\midrule
512 (5{,}120) & 72.7 & 33.5 & 35.5 & $0.46\times$ & $0.49\times$ \\
2{,}048 (20{,}480) & 72.7 & 182.0 & 109.8 & $2.5\times$ & $1.5\times$ \\
8{,}192 (81{,}920) & 72.7 & 2{,}030 & 872.5 & $27.9\times$ & $12.0\times$ \\
32{,}768 (327{,}680) & 72.7 & 29{,}965 & 11{,}309 & $411.9\times$ & $155.5\times$ \\
\bottomrule
\end{tabular}
\end{table}
\paragraph{Cost at the scale of the training scenes.} On the small scenes of the corpus the balance reverses (Tab.~\ref{tab:cost}, one A100): a single-view forward of \nmr{} takes 130\,ms against RenderFormer's 42--49\,ms, and for a training step it runs 1.09\,s against 0.72--0.91\,s. With $5\times$ fewer steps taken to 25\,dB, \nmr{} still reaches it in 4.8\,h of training against 15.9--20.3\,h.
\begin{table}[htbp]
\centering
\caption{Cost at the scale of the training scenes on one A100 80\,GB GPU (corpus scenes). Inference: median time per image on the held-out scenes, batch size 1, bf16 autocast as in Tab.~\ref{tab:scaling}; RenderFormer with SDPA / FlashAttention. Training step: median time per optimizer step with three training scenes from the corpus (the per-GPU share of the global batch of 24) on a single A100 GPU, each model in its training configuration: \nmr{} under bf16 autocast, RenderFormer with / without activation recomputation. Time to 25\,dB: steps to 25\,dB (Tab.~\ref{tab:teaser}) multiplied by per-step latency on an 8-GPU node without communication overhead.}
\label{tab:cost}
\footnotesize
\begin{tabular}{lcccc}
\toprule
Model & Inference (ms) & Training step (s) & Steps to 25\,dB & Time to 25\,dB (h) \\
\midrule
\nmr{} & 130 & 1.09 & 16k & 4.8 \\
RenderFormer & 42 / 49 & 0.91 / 0.72 & 80k & 15.9--20.3 \\
\bottomrule
\end{tabular}
\end{table}

\begin{table}[htbp]
\centering
\footnotesize
\caption{Single-forward inference time (ms, median) at $128^2$ and $256^2$ for the settings of Tab.~\ref{tab:scaling}. Speedups divide RenderFormer's time by \nmr{}'s.}
\label{tab:scaling-all}
\begin{tabular}{lrrrrr}
\toprule
& & \multicolumn{2}{c}{RenderFormer} & \multicolumn{2}{c}{Speedup} \\
\cmidrule(lr){3-4}\cmidrule(lr){5-6}
Objects & \nmr{} & sdpa & flash & sdpa & flash \\
\midrule
\multicolumn{6}{l}{$128^2$} \\
1   & 28.6  & 17.4     & 28.9    & $0.61\times$ & $1.01\times$ \\
10  & 30.3  & 25.4     & 33.1    & $0.84\times$ & $1.09\times$ \\
50  & 47.6  & 245.7    & 137.8   & $5.2\times$  & $2.9\times$ \\
100 & 80.8  & 823.6    & 384.7   & $10.2\times$ & $4.8\times$ \\
200 & 180.2  & 3{,}022.4 & 1{,}267.7 & $16.8\times$ & $7.0\times$ \\
500 & 722.7 & 18{,}236.4 & 6{,}959.2 & $25.2\times$ & $9.6\times$ \\
\midrule
\multicolumn{6}{l}{$256^2$} \\
1   & 31.9  & 17.4     & 28.9    & $0.55\times$ & $0.91\times$ \\
10  & 34.0  & 26.5     & 33.2    & $0.78\times$ & $0.98\times$ \\
50  & 52.7  & 246.5    & 138.3   & $4.7\times$  & $2.6\times$ \\
100 & 88.4  & 823.8    & 386.7   & $9.3\times$ & $4.4\times$ \\
200 & 191.2 & 3{,}023.4 & 1{,}269.5 & $15.8\times$ & $6.6\times$ \\
500 & 740.6 & 18{,}260.0 & 6{,}964.9 & $24.7\times$ & $9.4\times$ \\
\bottomrule
\end{tabular}
\end{table}
\begin{table}[t]
\centering
\caption{Token budget per object, on the ablation subset and held-out split of Tab.~\ref{tab:ablation} under its 100k-step schedule, on the model without invariant form-factor attention (Sec.~\ref{sec:method-inv}), mean over seeds 42--44: PSNR per-image (mean $\pm$ sample standard deviation) / pooled, LPIPS, and $\ell_1$ in log radiance. Step time is the relative training step latency normalized to $M{=}32$.}
\label{tab:tokens}
\small
\resizebox{\ifdim\width>\linewidth\linewidth\else\width\fi}{!}{%
\begin{tabular}{lcccc}
\toprule
Patch tokens per object $M$ & PSNR $\uparrow$ & LPIPS $\downarrow$ & $\ell_1(\log)$ $\downarrow$ & Step time \\
\midrule
8  & $29.45 \pm 0.02$ / \textbf{25.93} & \textbf{0.062} & 0.021 & $0.79\times$ \\
16 & $\mathbf{29.47} \pm 0.02$ / \textbf{25.93} & \textbf{0.062} & 0.021 & $0.83\times$ \\
32 (default) & $29.43 \pm 0.07$ / 25.92 & 0.063 & 0.021 & $1.00\times$ \\
64 & $29.44 \pm 0.06$ / 25.92 & 0.063 & 0.021 & $1.09\times$ \\
\bottomrule
\end{tabular}}
\end{table}
\begin{table}[t]
\centering
\caption{Batched throughput (ms per image, median) of the two 206M models on eight-object scenes (one H100 80\,GB GPU, bf16, five warm-up and twenty timed forward passes, host-to-device transfers excluded; 512 triangles per object for RenderFormer, 2{,}048 points per object for \nmr{}).}
\label{tab:throughput}
\small
\resizebox{\ifdim\width>\linewidth\linewidth\else\width\fi}{!}{%
\begin{tabular}{lcccccccc}
\toprule
& \multicolumn{3}{c}{$128^2$} & \multicolumn{3}{c}{$256^2$} & \multicolumn{2}{c}{$512^2$} \\
\cmidrule(lr){2-4}\cmidrule(lr){5-7}\cmidrule(lr){8-9}
Batch size & 1 & 8 & 32 & 1 & 8 & 32 & 1 & 8 \\
\midrule
RenderFormer (sdpa)  & \textbf{21.5} & 13.7 & 12.8 & \textbf{22.5} & 15.2 & 14.4 & \textbf{29.2} & 22.3 \\
RenderFormer (flash) & 30.6 & 12.4 & 10.7 & 30.4 & 13.5 & 12.1 & 33.0 & \textbf{19.4} \\
\nmr{} (sdpa, ours)  & 30.3 & \textbf{7.1} & \textbf{5.5} & 33.1 & \textbf{13.2} & \textbf{11.9} & 73.2 & 51.3 \\
\bottomrule
\end{tabular}}
\end{table}

\noindent \textbf{Batched throughput.} On small scenes the batched regime (Tab.~\ref{tab:throughput}) favors \nmr{} at $128^2$ and $256^2$ from batch size 8 on, where object-centric tokens allow efficient parallel processing; at $512^2$, ray tokens dominate per-image overhead. 

\noindent \textbf{Multi-view amortization.} Both models compute a view-independent scene representation once and replay only the view-dependent part per view. Tab.~\ref{tab:amortized} times the two parts separately: the split reproduces the output of the unmodified forward pass to within bf16 rounding, and the sum of the two parts matches one end-to-end forward within 0.9\%. \nmr{}'s view-independent part is the cheaper one from 50 objects on (38 versus 118\,ms under FlashAttention, 38 versus 227\,ms under the shared backend), and the gap grows with the object count as in Tab.~\ref{tab:scaling}. Its view-dependent part, however, costs 52--58\,ms at $512^2$, 2.1--3.3 times RenderFormer's view transformer, because it decodes 16{,}384 ray tokens through the DPT head. At $512^2$ the amortized per-view cost therefore favors RenderFormer on these 8- and 50-object scenes (16.0 and 29.6 versus 52.6 and 58.5\,ms per view at 64 views). At $256^2$ the view-dependent parts are comparable (13--21\,ms) and the view-independent part decides: the two models are within a millisecond per view at eight objects and \nmr{} is ahead at fifty (16.1 versus 22.4\,ms at 64 views).
\begin{table}[t]
\centering
\caption{Multi-view amortization of the two 206M models (one H100, bf16, batch size 1, random weights, synthetic scenes of 512 triangles per object, FlashAttention for both, three warm-up and ten timed iterations, medians). $t_{\text{scene}}$: view-independent part (\nmr{}: encoder and scene transformer; RenderFormer: triangle-token transformer); $t_{\text{view}}$: view-dependent part (\nmr{}: rasterization, ray tokens, predictor and DPT head; RenderFormer: camera-space triangle positions, ray generation, view transformer and DPT head). The per-view cost of $V$ views is $(t_{\text{scene}} + V\,t_{\text{view}})/V$. \nmr{}'s scene transformer computes its float32 attention through PyTorch SDPA under both backends. Under the shared sdpa backend $t_{\text{scene}}$ is 20.0/37.7\,ms for \nmr{} and 15.2/227\,ms for RenderFormer at 8/50 objects; $t_{\text{view}}$ changes by less than 1\,ms for \nmr{} and becomes 14.8/34.7\,ms at $512^2$ for RenderFormer.}
\label{tab:amortized}
\small
\resizebox{\ifdim\width>\linewidth\linewidth\else\width\fi}{!}{%
\begin{tabular}{llcccccc}
\toprule
Resolution & Model & Objects & $t_{\text{scene}}$ (ms) & $t_{\text{view}}$ (ms) & ms/view, $V{=}1$ & $V{=}16$ & $V{=}64$ \\
\midrule
\multirow{4}{*}{$512^2$} & \nmr{} & 8 & 19.5 & 52.3 & 71.8 & 53.5 & 52.6 \\
 & RenderFormer & 8 & 17.7 & 15.7 & 33.4 & 16.8 & 16.0 \\
 & \nmr{} & 50 & 37.7 & 57.9 & 95.7 & 60.3 & 58.5 \\
 & RenderFormer & 50 & 117.5 & 27.8 & 145.3 & 35.1 & 29.6 \\
\midrule
\multirow{4}{*}{$256^2$} & \nmr{} & 8 & 19.4 & 13.9 & 33.3 & 15.1 & 14.2 \\
 & RenderFormer & 8 & 17.5 & 13.3 & 30.9 & 14.4 & 13.6 \\
 & \nmr{} & 50 & 37.7 & 15.5 & 53.2 & 17.9 & 16.1 \\
 & RenderFormer & 50 & 117.6 & 20.5 & 138.1 & 27.9 & 22.4 \\
\bottomrule
\end{tabular}}
\end{table}

\section{Architecture}
\label{app:arch}
Tab.~\ref{tab:arch} lists the modules of \nmr{} with their parameter counts (205{,}942{,}972 in total). The scene transformer uses QK-LayerNorm, rotary frequencies $1, 5^{1/3}, 5^{2/3}, 5$ over the 12 descriptor coordinates (96 rotated channels, 32 pass-through, rotate-half convention), 16 register tokens positioned at the mean patch centroid, and the receiver/sender radiosity encodings of Sec.~\ref{sec:method-bias}. The predictor fuses the last three scene-transformer layers into its context, uses rotary frequencies $1, 3^{1/3}, 3^{2/3}, 3$ over the spherical ray/object coordinates, adds the routing bias with one learned gate per layer, and decodes the ray tokens of the $4\times4$ pixel patches with a DPT head (256 features, output channels 128/256/512/1{,}024) that also receives the 14 rasterized surface channels through a zero-initialized $1\times1$ convolution. 

RenderFormer (205.2M): the released Base configuration, 12 view-independent and 6 view-dependent layers, width 768, 6 heads, feed-forward 3{,}072, DPT decoder, $8\times8$ ray patches. 

RenderFormer-V2 (207.6M): the released 200M configuration, which orders tokens along a Hilbert curve and restricts attention among them to a sliding window of $\pm256$ tokens (FlexAttention) with attention sinks (16 register tokens and one summary token per 64 geometry tokens), uses a Swin-style view transformer, and encodes $32\times32$ material patches into $4\times4\times80$ VAE latents projected to the token width; frozen pretrained encoders of its released input pipeline (a material-to-latent mapper and an image VAE) produce these latents and that of the empty environment map.
\begin{table}[htbp]
\centering
\footnotesize
\caption{Modules of \nmr{} and their parameter counts, totaling 205{,}942{,}972 parameters as counted from a trained checkpoint. Widths and layer counts are those of Sec.~\ref{sec:method-overview}.}
\label{tab:arch}
\setlength{\aboverulesep}{0pt}
\setlength{\belowrulesep}{0pt}
\renewcommand{\arraystretch}{1.25}
\begin{tabularx}{\linewidth}{l>{\raggedright\arraybackslash}Xr}
\toprule
\textbf{Module} & \textbf{Configuration} & \textbf{Parameters} \\
\midrule
Point encoder & 16-D inputs, hidden 512/1{,}024/1{,}024, backbone 2{,}048, GroupNorm; 32 FPS patch tokens of 128 neighbors & 9{,}985{,}024 \\
\rowcolor{rowgray} Scene transformer & 12 layers, width 1{,}024, 8 heads of 128, FFN 1{,}890, QK-LayerNorm, OBB-RoPE & 96{,}962{,}716 \\
Register tokens & 16 & 16{,}384 \\
\rowcolor{rowgray} Receiver/sender encoders & two MLPs $8\to256\to1{,}024$ & 530{,}944 \\
Invariant-attention gains & two MLPs $8\to64\to384$ (Sec.~\ref{sec:method-inv}) & 51{,}072 \\
\rowcolor{rowgray} Object FiLM & intrinsic-complexity statistics $\to$ scale/shift & 2{,}112{,}512 \\
Ray encoder & CamRay map + 5 hit channels per pixel, $4\times4$ patches & 135{,}168 \\
\rowcolor{rowgray} Predictor & 6 cross+self-attention layers, width 1{,}024, FFN 1{,}496, spherical RoPE, routing gates, camera FiLM (2{,}108{,}416), DPT head & 96{,}149{,}152 \\
\midrule
Total & & 205{,}942{,}972 \\
\bottomrule
\end{tabularx}
\end{table}

\noindent \textbf{Implementation details omitted from Sec.~\ref{sec:method}.} OBB-RoPE: each patch token $i$ carries $\mathbf{x}^{(i)} = [\mathbf{c}^{(i)} \,\|\, \mathbf{e}_1^{(i)} \,\|\, \mathbf{e}_2^{(i)} \,\|\, \mathbf{e}_3^{(i)}] \in \mathbb{R}^{12}$, its patch centroid and the principal axes of its parent object's samples (covariance eigendecomposition) scaled by the standard deviations along them, shared by the object's tokens; each coordinate is mapped through $F{=}4$ logarithmically spaced frequencies to phases $\mathbf{a}^{(i)}$, and queries and keys are split into a 96-dimensional rotary block and a 32-dimensional pass-through block, $\tilde{\mathbf{q}} = [\mathbf{q}_r \odot \cos \mathbf{a}^{(i)} + \mathrm{rot}(\mathbf{q}_r) \odot \sin \mathbf{a}^{(i)} \,\|\, \mathbf{q}_p]$ with the rotate-half pairing $\mathrm{rot}(\cdot)$. 

Register tokens are positioned at the masked mean of the scene's patch centroids (with a small numerical epsilon). Eigendecompositions and camera transforms of the descriptor are evaluated in single precision; in the predictor, centroids receive the full rigid camera transform and axes the rotation only. The rotate-half convention pairs each channel of the first half of the rotary block with its partner in the second half, with a sign flip. 

The receiver/sender MLPs of the radiosity encodings (Sec.~\ref{sec:method-bias}) are $8\to256\to1{,}024$. The intrinsic-complexity FiLM reads covariance spectra, radial moments, and convexity of the object's samples. CamRay assumes square images with a centered principal point; the ray-token projection is followed by a LayerNorm. Patches without rasterized hits carry a depth coordinate of zero in the spherical frame.

\noindent \textbf{Invariant form-factor attention (Sec.~\ref{sec:method-inv}).} The receiver and sender gains are two MLPs $8\to64\to 12 \cdot 8 \cdot 4$ over the eight radiosity statistics of Sec.~\ref{sec:method-bias}, one gain per layer, head, and term; the last layer of the sender MLP is initialized at zero, so the invariant terms start at exactly zero. The 16 features per head are appended to the queries and keys after the rotary rotation, the values are padded with zeros, and the attention of these layers is computed in float32 with the softmax scale of the 128-dimensional heads. Centroids are centered at the mean of the scene's patch centroids and divided by their root-mean-square radius (Prop.~\ref{prop:inv}), which keeps the float32 expansion well conditioned; each mean normal averages, without normalization, the normals of the 128 samples nearest to the patch centroid, the neighborhood of the point encoder, so a patch whose normals cancel carries a short vector rather than an arbitrary direction; register tokens carry a zero normal. The gradients of the gain MLPs are clipped separately from the rest of the model. The extension adds 0.05M parameters (the two gain MLPs, $8 \to 64 \to 384$; Tab.~\ref{tab:arch}).

\section{Seed replicates of the ablations}
\label{app:seeds}
\begin{table}[htbp]
\centering
\caption{Ablation rows ($256^2$, 100k steps, mean over seeds 42--44; $^\dagger$seeds 42--47) with the metrics of Tab.~\ref{tab:ablation}. PSNR per-image (mean $\pm$ sample standard deviation) / pooled; $\Delta$: per-image change from the full model, paired within each seed ($^\dagger$seeds 45--47 against the full model's three-seed mean). Below the rule: two controls of invariant form-factor attention (Sec.~\ref{sec:method-inv}).}
\label{tab:ablation-full}
\resizebox{\ifdim\width>\linewidth\linewidth\else\width\fi}{!}{%
\footnotesize
\setlength{\tabcolsep}{5pt}%
\begin{tabular}{>{\raggedright\arraybackslash}p{5.3cm}cccc}
\toprule
Variant & PSNR $\uparrow$ & LPIPS $\downarrow$ & $\ell_1(\log)$ $\downarrow$ & $\Delta$ \\
\midrule
Full model & $\mathbf{29.96} \pm 0.01$ / \textbf{26.16} & \textbf{0.059} & \textbf{0.019} & --- \\
No invariant form-factor attention (\S\ref{sec:method-inv}) & $29.43 \pm 0.07$ / 25.92 & 0.063 & 0.021 & $-0.52 \pm 0.08$ \\
No inter-object scene attention (\S\ref{sec:method-overview})$^\dagger$ & $27.50 \pm 0.66$ / 24.89 & 0.072 & 0.036 & $-2.46 \pm 0.66$ \\
No FiLM conditioning (\S\ref{sec:method-film}) & $29.81 \pm 0.02$ / 26.09 & 0.060 & \textbf{0.019} & $-0.15 \pm 0.02$ \\
No register tokens & $29.82 \pm 0.02$ / 26.10 & 0.060 & \textbf{0.019} & $-0.13 \pm 0.01$ \\
No rasterizer channels (hit geometry, routing bias) & $26.84 \pm 0.07$ / 24.41 & 0.092 & 0.025 & $-3.12 \pm 0.08$ \\
\midrule
Invariant terms frozen at zero (float32 attention kept) & $29.43 \pm 0.03$ / 25.92 & 0.063 & 0.021 & $-0.53 \pm 0.02$ \\
No OBB-RoPE (invariant attention kept) & $29.87 \pm 0.04$ / 26.12 & \textbf{0.059} & \textbf{0.019} & $-0.08 \pm 0.04$ \\
\bottomrule
\end{tabular}}
\end{table}

\noindent \textbf{Rasterizer controls.} Tab.~\ref{tab:raster} separates the rasterizer channels from the scene representation under the 100k-step schedule of Tab.~\ref{tab:ablation-full}, over seeds 42--44, on the model without invariant form-factor attention (row ``No invariant form-factor attention'' of Tab.~\ref{tab:ablation-full}). Removing the rasterized splat into the output head leaves PSNR unchanged ($+0.01 \pm 0.07$\,dB). Removing the scene tokens from the view decoder while keeping every rasterizer channel and the splat costs $5.66 \pm 0.06$\,dB and triples the log-radiance error ($0.062$ against $0.021$), more than removing the rasterizer channels ($3.12 \pm 0.05$\,dB). Removing the rasterizer channels and the splat together costs $3.09 \pm 0.20$\,dB. The rasterizer channels identify the surface each ray meets, and the scene tokens provide the global 3D scene context needed to shade it: neither suffices alone.

\begin{table}[htbp]
\centering
\caption{Rasterizer controls under the 100k-step schedule of Tab.~\ref{tab:ablation-full} (mean over seeds 42--44), on the model without invariant form-factor attention. Columns indicate whether the view decoder receives the scene tokens, the rasterizer channels (hit channels and routing bias), and the rasterized splat into the output head. PSNR per-image (mean $\pm$ sample standard deviation) / pooled; $\Delta$: per-image change from the reference row, paired within each seed.}
\label{tab:raster}
\footnotesize
\resizebox{\ifdim\width>\linewidth\linewidth\else\width\fi}{!}{%
\setlength{\tabcolsep}{4pt}%
\begin{tabular}{>{\raggedright\arraybackslash}p{2.6cm}ccccccc}
\toprule
Variant & \shortstack{Scene\\tokens} & \shortstack{Rasterizer\\channels} & Splat & \shortstack{PSNR\\$\uparrow$} & \shortstack{LPIPS\\$\downarrow$} & \shortstack{$\ell_1(\log)$\\$\downarrow$} & $\Delta$ \\
\midrule
Reference (no invariant attention) & \cmark & \cmark & \cmark & $29.43 \pm 0.07$ / 25.92 & 0.063 & 0.021 & --- \\
No rasterized splat & \cmark & \cmark & \xmark & $29.44 \pm 0.04$ / 25.92 & 0.062 & 0.021 & $+0.01 \pm 0.07$ \\
No rasterizer channels & \cmark & \xmark & \cmark & $26.31 \pm 0.09$ / 24.08 & 0.099 & 0.027 & $-3.12 \pm 0.05$ \\
No rasterizer channels or splat & \cmark & \xmark & \xmark & $26.34 \pm 0.13$ / 24.01 & 0.094 & 0.027 & $-3.09 \pm 0.20$ \\
No scene tokens & \xmark & \cmark & \cmark & $23.77 \pm 0.01$ / 21.44 & 0.109 & 0.062 & $-5.66 \pm 0.06$ \\
\bottomrule
\end{tabular}}
\end{table}

Tab.~\ref{tab:ablation-full} reports the mean of seeds 42, 43, and 44 (42--47 without inter-object attention) under the same 100k-step schedule, and Tab.~\ref{tab:seeds} lists every seed.
Each replicate changes both the initialization and the data order; all runs are scored by the same evaluator on the same
2{,}948 held-out images. The full model's seed-to-seed standard deviation is $0.01$\,dB. Paired within seeds, removing the rasterizer channels costs
$3.12 \pm 0.08$\,dB and removing invariant form-factor attention $0.52 \pm 0.08$\,dB;
removing FiLM costs $0.15 \pm 0.02$\,dB and removing the register tokens costs $0.13 \pm 0.01$\,dB, with the same sign in every seed.
Without inter-object attention the loss depends on the seed, from $1.41$ to $3.03$\,dB, so this row reports seeds 42--47; seeds 45--47 have no full-model run of their own and are compared with its three-seed mean. The rasterized splat
into the output head (Sec.~\ref{sec:method-rays}) is kept in every row. The result is thus carried by the rasterizer channels, the attention between objects, and invariant form-factor attention, not by the auxiliary conditioning.
\begin{table}[htbp]
\centering
\footnotesize
\caption{Per-image PSNR (dB) of every row of Tab.~\ref{tab:ablation-full} at 100k steps (the schedule of Tab.~\ref{tab:ablation}) for seeds 42, 43,
and 44 (42--47 without inter-object attention), with the mean and sample standard deviation. $\Delta$ is paired within each seed, against the full model; seeds 45--47 against its three-seed mean. Below the rule: two controls of invariant form-factor attention (Sec.~\ref{sec:method-inv}).}
\label{tab:seeds}
\resizebox{\ifdim\width>\linewidth\linewidth\else\width\fi}{!}{%
\footnotesize
\setlength{\tabcolsep}{4pt}%
\begin{tabular}{>{\raggedright\arraybackslash}p{3.4cm}cccccccc}
\toprule
& \multicolumn{6}{c}{Seed} & & \\
\cmidrule(lr){2-7}
Variant & 42 & 43 & 44 & 45 & 46 & 47 & Mean $\pm$ SD & $\Delta$ \\
\midrule
Full model & 29.97 & 29.95 & 29.96 & -- & -- & -- & $29.96 \pm 0.01$ & --- \\
No invariant form-factor attention & 29.36 & 29.49 & 29.45 & -- & -- & -- & $29.43 \pm 0.07$ & $-0.52 \pm 0.08$ \\
No inter-object scene attention$^\dagger$ & 28.55 & 26.95 & 26.93 & 27.69 & 26.98 & 27.89 & $27.50 \pm 0.66$ & $-2.46 \pm 0.66$ \\
No FiLM conditioning & 29.79 & 29.81 & 29.82 & -- & -- & -- & $29.81 \pm 0.02$ & $-0.15 \pm 0.02$ \\
No register tokens & 29.84 & 29.81 & 29.82 & -- & -- & -- & $29.82 \pm 0.02$ & $-0.13 \pm 0.01$ \\
No rasterizer channels & 26.76 & 26.88 & 26.88 & -- & -- & -- & $26.84 \pm 0.07$ & $-3.12 \pm 0.08$ \\
\midrule
Invariant terms frozen at zero & 29.43 & 29.41 & 29.46 & -- & -- & -- & $29.43 \pm 0.03$ & $-0.53 \pm 0.02$ \\
No OBB-RoPE & 29.83 & 29.88 & 29.90 & -- & -- & -- & $29.87 \pm 0.04$ & $-0.08 \pm 0.04$ \\
\bottomrule
\end{tabular}}
\end{table}

\noindent \textbf{Inter-object attention.} Restricting the self-attention of the scene transformer to the tokens of each object (the register tokens form a block of their own, so nothing relays between objects; the view decoder still cross-attends to every object's tokens, emitters included) lowers per-image PSNR by $2.46 \pm 0.66$\,dB over seeds 42--47 (pooled $1.27 \pm 0.36$\,dB; LPIPS $+23\%$, $\ell_1(\log)$ $+91\%$). The loss depends on the seed: $1.41$, $2.07$, and $2.27$\,dB for seeds 42, 47, and 45, and $2.98$ to $3.03$\,dB for seeds 43, 44, and 46. The drop is present at every scene size and largest in the sparsest scenes: $2.87$, $2.47$, $2.35$, and $2.23$\,dB for at most 7, 8--10, 11--13, and at least 14 entities (seeds 42--44, paired per image within each seed), where the walls and floor lit by the emitters fill more of the image.

\noindent \textbf{Invariant form-factor attention.} Trained from scratch under the protocol of Tab.~\ref{tab:ablation}, the full model
reaches $29.96 \pm 0.01$\,dB per-image PSNR against $29.43 \pm 0.07$\,dB without the invariant channels of Sec.~\ref{sec:method-inv}, a paired difference of $0.52 \pm 0.08$\,dB over seeds 42--44 (pooled $26.16$ against $25.92$\,dB, LPIPS $0.059$ against $0.063$, $\ell_1(\log)$ $0.019$ against $0.021$; three-seed means). Two controls isolate the
mechanism. The invariant layers compute attention in float32, the rest of the model in bf16; the full model with the gains frozen at their initialization, so that the invariant terms are exactly zero, reaches $29.43 \pm 0.03$\,dB over seeds 42--44, $0.53 \pm 0.02$\,dB below the full model and level with the model without the invariant channels ($-0.00 \pm 0.08$\,dB paired): precision does not account for the gain. Without OBB-RoPE (no rotary encoding in the scene transformer, invariant channels kept) the model reaches $29.87 \pm 0.04$\,dB, $0.08 \pm 0.04$\,dB below the full model and $0.44 \pm 0.04$\,dB above the model without the invariant channels (paired), so the invariants carry more than the rotary encoding.

\noindent \textbf{Controls.} On the model without invariant form-factor attention, varying the token budget per object over $M \in \{8, 16, 32, 64\}$ (Tab.~\ref{tab:tokens}) moves the
three-seed mean by at most $0.04$\,dB from the default $M{=}32$, so the fixed budget is not the binding constraint at this scale.

\noindent \textbf{Pooling RenderFormer's triangles per object.} \nmr{} represents every object by a fixed budget of 32 patch tokens. To test whether this budget alone explains the difference from RenderFormer, we give RenderFormer the same budget: before global transport, a learned pooler maps the triangle tokens of each object to 32 tokens (learned queries at 32 farthest-point anchors per object and two layers of RenderFormer's own transformer decoder; ${+}11.6\%$ parameters), and both of its transformers then run on the pooled tokens instead of the triangles. With the authors' released training loop and the 100k-step schedule of Tab.~\ref{tab:ablation-full}, pooling lowers RenderFormer's per-image PSNR from $23.56 \pm 0.21$ to $23.19 \pm 0.15$\,dB, by $0.38 \pm 0.09$\,dB paired within seeds (Tab.~\ref{tab:rf-pool}), while \nmr{} reaches $29.96 \pm 0.01$\,dB under the same schedule. 

The difference between the two architectures is therefore not a matter of the token budget per object. Within \nmr{}, the ablations above attribute the quality to the rasterizer channels and the attention between objects. In the pooled model the view-dependent stage no longer attends to individual triangles, which may account for part of the loss. \nmr{} provides the primary visibility through its rasterizer channels instead.

\begin{table}[htbp]
\centering
\footnotesize
\caption{RenderFormer with the triangle tokens of each object pooled into 32 tokens before global transport, trained with the authors' released training loop under the 100k-step schedule of Tab.~\ref{tab:ablation-full} (mean over seeds 42--44). PSNR per-image (mean $\pm$ sample standard deviation) / pooled; $\Delta$: per-image change from RenderFormer without pooling, paired within each seed. The last row is the full model of Tab.~\ref{tab:ablation-full}.}
\label{tab:rf-pool}
\setlength{\tabcolsep}{4pt}
\begin{tabular}{lcccc}
\toprule
Model & PSNR $\uparrow$ & LPIPS $\downarrow$ & $\ell_1(\log)$ $\downarrow$ & $\Delta$ \\
\midrule
RenderFormer & $23.56 \pm 0.21$ / 21.84 & 0.166 & 0.041 & --- \\
RenderFormer, 32 pooled tokens per object & $23.19 \pm 0.15$ / 21.59 & 0.165 & 0.041 & $-0.38 \pm 0.09$ \\
\midrule
\nmr{} (full model) & $29.96 \pm 0.01$ / 26.16 & 0.059 & 0.019 & --- \\
\bottomrule
\end{tabular}
\end{table}

\paragraph{Uncertainty of the main comparison.} Paired per-image PSNR differences at matched step 600k, with 95\% confidence intervals from a scene-level bootstrap (the four views of a held-out scene are resampled together; 10{,}000 resamples): \nmr{} leads RenderFormer by $+3.26$\,dB ($[+3.12, +3.39]$) and RenderFormer-V2 by $+7.63$\,dB ($[+7.45, +7.81]$), and it is better on 90.1\% and 98.1\% of the 4{,}112 held-out images (92.1\% and 98.2\% of the 1{,}028 scenes). Each model is trained once (seed 42), so these intervals cover the sampling of held-out scenes, not training variance; under the ablation protocol, the sample standard deviation of per-image PSNR across seeds 42--44 is $0.01$\,dB for \nmr{} and $0.21$\,dB for RenderFormer (Tabs.~\ref{tab:ablation} and~\ref{tab:rf-pool}).

\section{Design trade-offs}
\label{app:tradeoffs}
Tab.~\ref{tab:compare} places \nmr{} among feed-forward renderers of explicit scenes.

\begin{table}[htbp]
\centering
\caption{Feed-forward renderers of explicit scenes: geometric input, what the sequence length is proportional to ($T$ triangles, $N$ objects, $M$ learned queries per object), full global illumination without per-scene optimization (GI), and a per-scene representation reused across views (Amortized).}
\label{tab:compare}
\footnotesize\setlength{\tabcolsep}{4pt}
\begin{tabular}{@{}>{\raggedright\arraybackslash}p{4.2cm}>{\raggedright\arraybackslash}p{3.3cm}>{\raggedright\arraybackslash}p{2.6cm}cc@{}}
\toprule
Method & Input & Tokens & GI & Amortized \\
\midrule
Per-scene fields \citep{mildenhall2020nerf, kerbl2023gaussians} & images & --- & per scene & --- \\
RenderFormer \citep{zeng2025renderformer} & triangles & $O(T)$ & \checkmark & \checkmark \\
RenderFormer-V2 \citep{zeng2026renderformerv2} & triangles, media, environment maps & $O(T)$, windowed & \checkmark & \checkmark \\
RenderFormer++ \citep{du2026renderformerpp} & triangles & $O(T_i)$ per object, then $O(NM)$ & \checkmark & \checkmark \\
\nmr{} (ours) & point inputs (from meshes or point clouds) & $O(N)$, 32 per object & \checkmark & \checkmark \\
\bottomrule
\end{tabular}
\end{table}
\noindent \textbf{Token budget per object.} The fixed budget is what makes the cost independent of mesh resolution and what the mechanisms of Sec.~\ref{sec:method} compensate for. The ablation shows that the compensation is uneven: the rasterizer channels, which tell the decoder where each pixel lands, recover three decibels (Tab.~\ref{tab:ablation}), whereas removing OBB-RoPE from the scene transformer costs $0.08 \pm 0.04$\,dB (Tab.~\ref{tab:ablation-full}); and Tab.~\ref{tab:tokens} shows, on the model without invariant form-factor attention, that the budget itself can be varied eightfold within $0.04$\,dB (29.45 / 29.47 / 29.43 / 29.44\,dB for $M{=}8/16/32/64$), with 16 tokens matching 32 at 17\% lower step time. Objects whose appearance varies strongly across their surface remain the hard case, and a variable token budget per object is the natural extension.

\noindent \textbf{Attention backend.} The two models favor different attention kernels: RenderFormer's dense triangle attention gains $2.6\times$ from FlashAttention at 500 objects (Tab.~\ref{tab:scaling}), whereas \nmr{}'s routed cross-attention materializes a ray-by-token bias and therefore runs through PyTorch's scaled-dot-product path, as does the float32 attention of its scene transformer (Sec.~\ref{sec:method-inv}). Efficiency comparisons report both the shared backend and RenderFormer's fastest one.

\noindent \textbf{Disentangling 3D transport, explicit visibility, and decoder capacity.}
To determine whether \nmr{}'s gains stem from structured 3D light transport modeling, explicit surface visibility from the point rasterizer, or view decoder capacity (16k versus 4k ray tokens, both decoded by a DPT head), we address three questions:
\begin{itemize}
\item \textbf{3D light transport vs.\ explicit visibility:} While the point rasterizer provides local surface grounding (depth, normals, materials, instance mask), it contains no indirect illumination or shadows. In the control without scene tokens (Tab.~\ref{tab:raster}, on the model without invariant form-factor attention), the view decoder receives all rasterizer channels and ray tokens but no 3D scene tokens, and PSNR drops by $5.66 \pm 0.06$\,dB. Conversely, holding the rasterizer and decoder fixed, restricting the 3D scene transformer to attention within objects lowers PSNR by $2.46 \pm 0.66$\,dB and removing invariant form-factor attention by $0.52 \pm 0.08$\,dB (Tab.~\ref{tab:ablation}). Thus, with the rasterizer and decoder fixed, variants of the 3D scene transformer span $2.5$\,dB ($27.50$ to $29.96$\,dB).
\item \textbf{Decoder capacity and ray token density:} At $512^2$, \nmr{} uses $16{,}384$ ray tokens ($4\times4$ patches) and RenderFormer uses $4{,}096$ tokens ($8\times8$ patches), both with a DPT decoder, at matched total parameter counts ($\approx$206M). The control without the rasterized splat (Tab.~\ref{tab:raster}, same model) removes the per-pixel feature injection into the DPT head and leaves PSNR unchanged, so late convolutional upsampling does not bypass cross-attention transport. Training \nmr{} (without invariant form-factor attention) with RenderFormer's $8\times8$ ray patches under the ablation protocol (Tab.~\ref{tab:ablation}) lowers per-image PSNR by $0.60 \pm 0.09$\,dB, paired over seeds 42--44 ($28.83 \pm 0.03$ against $29.43 \pm 0.07$\,dB): the denser ray tokens contribute, but less than the attention between objects in the scene transformer, whose removal costs this model $1.59 \pm 0.25$\,dB over seeds 42--44 (the full model $2.46 \pm 0.66$\,dB, Tab.~\ref{tab:ablation}).
\item \textbf{Decoupling visibility from transport:} Integrating rasterized point channels directly into triangle transformers would fundamentally alter their input representation and cross-attention formulation away from mesh primitives. The controls of Tab.~\ref{tab:raster} isolate the role of rasterized guidance within the point-based paradigm, demonstrating that global 3D scene context remains essential.
\end{itemize}

\section{Unseen assets and larger scenes}
\label{app:unseen}
The held-out split shares the training meshes and the training range of scene sizes (Appendix~\ref{app:data}). To test generalization beyond both, we build a separate test set with the same generator and 500 Objaverse objects that never occur in training. The curation criteria admit no mesh beyond the 18{,}085 of the training pool, so the new objects are Objaverse-LVIS objects that meet every curation criterion except the Sketchfab-category filter; none matches a training mesh by identifier, file hash, or geometry. Three subsets are rendered and converted exactly like the corpus (four views per scene at $512^2$): \emph{A}, 250 scenes from the unchanged generator (3--17 entities, mean 9.7, against 10.3 on the held-out split); \emph{B}, 280 scenes with 25--50 entities, whose $n$ objects are scaled by $(3.10/n)^{1/3}$ so that they fill the room as in an average training scene; and \emph{C}, 250 scenes with the training object count and the object sizes of \emph{B}, which separates object size from entity count. Every \emph{B} scene exceeds the 8{,}192-triangle limit of the RenderFormer data (10{,}153--44{,}200 triangles after the same simplification, median 27{,}766); RenderFormer and RenderFormer-V2 receive these scenes with the limit raised for this evaluation only, so all three models run beyond their training range. Tab.~\ref{tab:unseen} reports the step-600k checkpoints with the metrics of Sec.~\ref{sec:exp-setup}; confidence intervals come from a scene-level bootstrap.

On unseen assets at the training scene size (\emph{A}), each model scores $+0.32$ to $+0.41$\,dB relative to its held-out result, so the new assets cost no accuracy. With 25--50 entities (\emph{B}), all three models lose accuracy ($B - A$: $-7.3$, $-6.9$, and $-4.8$\,dB for \nmr{}, RenderFormer, and RenderFormer-V2), and \nmr{} still leads RenderFormer by $+2.81$\,dB (95\% CI $[+2.63, +3.00]$) and RenderFormer-V2 by $+5.26$\,dB ($[+5.03, +5.49]$). Smaller objects alone (\emph{C}) are easier for all three models ($C - A$: $+2.4$, $+3.0$, and $+1.5$\,dB), so the drop on \emph{B} comes from the number of entities rather than from object size. \nmr{} leads both baselines on every subset: by $+3.23$ and $+7.69$\,dB on \emph{A} and by $+2.61$ and $+8.52$\,dB on \emph{C}.

\section{Imperfect geometry and robustness}
\label{app:partial}
Raw geometry acquired by physical depth sensors is frequently incomplete due to occlusions. To assess how \nmr{} behaves under unobserved surface regions, we evaluate two held-out scenes (five foreground objects each) where each object retains only the point inputs visible from a single simulated depth camera at the first view: points that are camera-facing, within the viewing frustum, and pass a z-buffer test against entity point splats (radius 2 pixels at $512^2$). Across objects, 28\%--48\% of the point inputs survive this visibility test. Resampling these to 2{,}048 points per object preserves the standard input budget while covering only the observed portions of each object; room surfaces and emitters remain complete.

\noindent \textbf{Zero-shot vs.\ fine-tuning protocol.}
We evaluate \nmr{} at the 600k-step checkpoint in two regimes (Tab.~\ref{tab:partial}), on the four original views of each scene and on eight new views per scene, path traced like the corpus from cameras drawn from the training camera distribution and never used for fitting. (1) \emph{Zero-shot inference}: without fine-tuning, the partial scans reach 26.33\,dB on the new views (26.76\,dB on the original views), against 34.24\,dB from complete geometry; unscanned surfaces render as voids. (2) \emph{Per-scene adaptation}: fine-tuning on the four original views for 400 steps (batch size 2, AdamW, learning rate $10^{-5}$, using the loss in Sec.~\ref{sec:method-training}) brings the partial scans to 37.51\,dB on those views, 1.51\,dB below the same fine-tuning from complete geometry (39.02\,dB), but to 30.30\,dB on the new views, and the fine-tuning from complete geometry does not carry over to the new views (33.58\,dB against 34.24\,dB without fine-tuning). The gains on the fitted views therefore mostly reflect fitting those views; on views that were not fitted, fine-tuning recovers 50\% of the 7.91\,dB lost to the missing surfaces. Fig.~\ref{fig:partial} shows three of the new views per scene.

\noindent\textbf{Zero-shot rendering from corrupted inputs.} Tab.~\ref{tab:robust} evaluates the 600k-step checkpoint without fine-tuning across various geometric corruptions on the first 1{,}028 held-out images. When corruptions are restricted to foreground objects (room and emitters clean), reducing point density to a quarter (512 samples per entity) or adding Gaussian position noise with $\sigma = 0.01$ of the object bounding-box diagonal leaves \nmr{} at 27.1 and 28.5\,dB, against RenderFormer's 28.7\,dB on clean input on the same images. Corrupting every entity (including high-coverage walls and floor) leads to steeper degradation because rasterized hit channels directly propagate geometric noise into the view decoder.
\begin{table}[htbp]
\centering
\caption{Zero-shot rendering from corrupted point inputs (no fine-tuning; \nmr{} at step 600k; first 1{,}028 held-out images; PSNR per-image / LPIPS). \emph{Objects}: only the foreground objects are corrupted; \emph{All}: every entity, including walls, floor, and emitter geometry (emission values unchanged). On clean input the same images score 32.1 (\nmr{}), 28.7 (RenderFormer), and 24.5\,dB (RenderFormer-V2), all at step 600k.}
\label{tab:robust}
\footnotesize
\begin{tabular}{llcc}
\toprule
Corruption & Level & Objects & All \\
\midrule
Clean input & --- & \multicolumn{2}{c}{32.1 / 0.043} \\
Point density (samples per entity) & 1{,}024 & 31.1 / 0.050 & 28.3 / 0.055 \\
 & 512 & 27.1 / 0.079 & 25.2 / 0.082 \\
 & 256 & 23.6 / 0.119 & 20.0 / 0.194 \\
Position noise ($\sigma$, \% of box diagonal) & 0.5 & 30.8 / 0.049 & 23.5 / 0.078 \\
 & 1 & 28.5 / 0.062 & 19.4 / 0.144 \\
 & 2 & 25.4 / 0.095 & 15.8 / 0.266 \\
Normal noise (mean angular error) & 14$^\circ$ & 31.4 / 0.054 & 25.0 / 0.133 \\
 & 35$^\circ$ & 29.7 / 0.076 & 19.7 / 0.309 \\
 & 57$^\circ$ & --- & 15.7 / 0.376 \\
Material noise (relative $\sigma$) & 10\% & --- & 30.6 / 0.056 \\
 & 25\% & --- & 27.5 / 0.090 \\
\bottomrule
\end{tabular}
\end{table}

\section{Additional related work}
\label{app:related}
\noindent \textbf{Hierarchical and meshless transport.} Hierarchical radiosity \citep{hanrahan1991rapid} decomposes the form-factor matrix of $n$ elements into $O(n)$ blocks by subdividing patches adaptively, and meshless hierarchical bases \citep{lehtinen2008meshless} represent illumination independently of the geometric discretization for precomputed direct-to-indirect transport. Both compute transport for a given scene while \nmr{} learns one mapping and applies it to unseen scenes.

\noindent \textbf{Neural representations of global illumination.} Neural radiance caching \citep{mueller2021nrc} adapts a network online inside a path tracer instead of pretraining it; active exploration \citep{diolatzis2022active} trains a neural generator for one variable scene given an explicit parameterization of its configuration; neural vertex features \citep{su2025vertex} store learnable features at mesh vertices, with an emphasis on neural radiosity; and object-centric feature grids model multi-frequency reflections in dynamic scenes \citep{mo2025dualband}. \nmr{} is trained once across a million scenes and applied to unseen scenes without further training.

\noindent \textbf{Point-based rendering and visibility.} Surface splatting \citep{zwicker2001surface} renders point clouds without connectivity using a screen-space elliptical weighted average filter. The point splat of \nmr{} is a simpler depth-tested splat with $7\times7$ dilation that supplies the primary G-buffer. Pointersect \citep{chang2023pointersect} infers the intersection of a ray with the surface underlying a point cloud with a set transformer, which enables ray tracing of point clouds, including global illumination. In contrast, \nmr{} traces no rays and predicts all lighting in one forward pass.

\section{Limitations, future work, and broader impact}
\label{app:impact}
Beyond the limitations of Sec.~\ref{sec:conclusion}, the baselines are trained under our matched capacity and budget rather than their authors' schedules; the fixed sample and token budget per object bounds the fine geometric detail that can be represented; and small scenes at batch size 1 favor triangle baselines before quadratic scaling dominates (Appendix~\ref{app:runtime}). This work uses only synthetic scenes assembled from publicly available Objaverse assets, involves no human subjects or personal data, and its intended use is faster physically based rendering; we do not foresee misuse specific to this method beyond that of rendering in general. The licenses of the assets, including the non-commercial terms of 1{,}093 of the 18{,}085 training meshes, are listed in the ethics statement.